\documentclass[10pt,twocolumn,letterpaper]{article}

\usepackage{wacv}              

\usepackage{graphicx}
\usepackage{amsmath}
\usepackage{amssymb}
\usepackage{booktabs}

\usepackage[pagebackref,breaklinks,colorlinks]{hyperref}

\usepackage[capitalize]{cleveref}
\crefname{section}{Sec.}{Secs.}
\Crefname{section}{Section}{Sections}
\Crefname{table}{Table}{Tables}
\crefname{table}{Tab.}{Tabs.}

\def\wacvPaperID{1278} 
\def\confName{WACV}
\def\confYear{2024}

\begin{document}

\title{Unseen Object Reasoning with Shared Appearance Cues}

\author{Paridhi Singh\\
{\tt\small paridhi@ridecell.com}
\and
Arun Kumar\\
{\tt\small arun.kumar@ridecell.com}
}
\maketitle

\begin{abstract}
    This paper introduces an innovative approach to open world recognition (OWR), where we leverage knowledge acquired from known objects to address the recognition of previously unseen objects. The traditional method of object modeling relies on supervised learning with strict closed-set assumptions, presupposing that objects encountered during inference are already known at the training phase. However, this assumption proves inadequate for real-world scenarios due to the impracticality of accounting for the immense diversity of objects. Our hypothesis posits that object appearances can be represented as collections of "shareable" mid-level features, arranged in constellations to form object instances. By adopting this framework, we can efficiently dissect and represent both known and unknown objects in terms of their appearance cues. Our paper introduces a straightforward yet elegant method for modeling novel or unseen objects, utilizing established appearance cues and accounting for inherent uncertainties. This representation not only enables the detection of out-of-distribution objects or novel categories among unseen objects but also facilitates a deeper level of reasoning, empowering the identification of the superclass to which an unknown instance belongs. This novel approach holds promise for advancing open world recognition in diverse applications.

\end{abstract}

\section{Introduction}
\label{sec:intro}
\begin{figure}
     \includegraphics[width=0.53\textwidth]{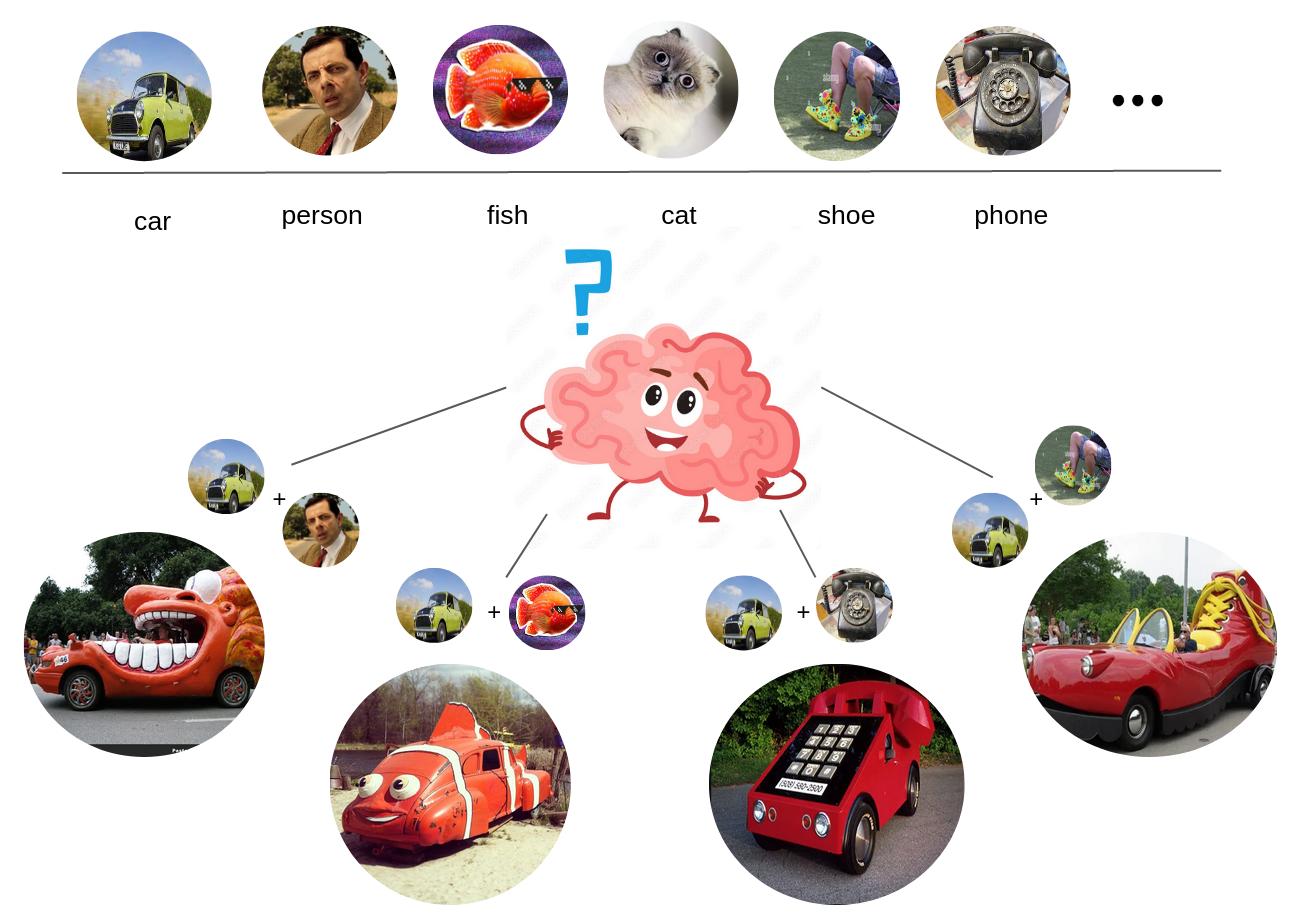}
     \caption{If human brains can successfully reason novel objects, why let the networks fail?}
     \label{fig:my_label}
 \end{figure}
In our paper, we present an innovative approach to open world recognition (OWR) that redefines how machines perceive and reason about objects. The traditional paradigm of supervised learning relies on closed world assumptions, where unfamiliar classes encountered during testing are dismissed as irrelevant to the trained model. Yet, in real-world scenarios like autonomous driving, infrequent and novel objects emerge regularly, defying closed world constraints.

Consider the case of an autonomous vehicle encountering a "construction vehicle." Conventional supervised models struggle to make sense of this unfamiliar entity, leading to detection failures and an inability to reason about it. Our method transcends these limitations by embracing a human-like approach to reasoning. Humans don't perceive objects as isolated labels but rather as interconnected appearances in a continuous space. We leverage this insight to redefine object appearances as constellations of "common" and "shareable" features, allowing us to flexibly represent known and unknown objects alike.

Central to our approach is a dynamic cost function that discerns whether a test-time object belongs to a known or unknown class. When facing an unknown object, we harness the web of feature similarities to infer its constitution from familiar classes. To illustrate, our method would enable the autonomous vehicle to deduce that the "construction vehicle" belongs to the superclass "vehicle," facilitating nuanced responses such as estimating size, predicting motion, and adapting navigation strategies.

The crux of our innovation lies in challenging the traditional learning paradigm, which aims to maximize dissimilarities between object classes. Instead, we introduce a paradigm shift that emphasizes the importance of capturing similarities, thereby enabling models to reason about objects in an interconnected manner. Through a simple yet elegant constellation-based representation, we reshape the landscape of open world recognition, addressing the limitations of closed world assumptions and empowering machines to comprehend and navigate the ever-evolving real world.
\section{Related Work}

Open Set Recognition (OSR) addresses the challenge of classifying instances from unseen or unknown categories, which were absent during training. Early work, such as that by Bendale and Boult~\cite{bendale2016towards}, laid the groundwork for OSR. Traditional training methods involve cross-entropy on known classes, with subsequent softmax predictions on query images.

To enhance OSR capabilities, researchers have explored diverse approaches. Generative Adversarial Networks (GANs) have played a role, as exemplified by studies like Neal et al.~\cite{neal2018open} and Ge et al.~\cite{ge2017generative}. An alternative perspective involves harnessing reconstruction error to identify open-set samples, as seen in methods like C2AE~\cite{oza2019c2ae} and Yoshihashi et al.~\cite{yoshihashi2019classification}.

The open-set problem closely intersects with the out-of-distribution challenge, where distinguishing unfamiliar instances is crucial. Approaches like Generalized ODIN~\cite{hsu2020generalized} and Enhancing OOD Detection~\cite{liang2017enhancing} contribute to this aspect.

Recent advances have seen the emergence of out-of-distribution-based strategies for open-set issues. Chen et al.~\cite{chen2021adversarial, chen2020learning} illustrate such methods. Notably, Vaze et al.~\cite{vaze2021open} exhibit the effectiveness of enhancing closed-set classifiers.

A significant subtask within OSR is the detection of novel categories. Han et al.~\cite{han2019learning} focus on this challenge, while Jia et al.~\cite{jia2021joint} propose a multimodal contrastive learning approach for end-to-end category discovery. Fini et al.'s UNO~\cite{fini2021unified} leverages labeled and unlabeled data jointly through label pseudo-label swapping.

A closely related work to ours is the method proposed by Vaze et al.~\cite{vaze2021open}, which employs clustering techniques to discern whether an unseen object is out-of-distribution. We extend this concept by not only identifying novel objects but also decomposing them into known and unknown appearances using information from learned object classes or supervision. Our hypothesis posits that the visual world is characterized by generalizable mid-level appearance cues, enabling us to effectively model or reason about unknown object classes with sufficient training data and diverse annotations.

In this paper, (1) we highlight the drawbacks with conventional supervision based methods, (2) point out how almost all of the open world recognition works are only out-of-distribution estimation techniques, (3) propose a novel way of representing objects that not only allows discovery of out-of-distribution, but also further reasons it, (4) a formulation and a proof-of-concept demonstrating how any unseen objects can be reasoned using information gained from known objects to regress the superclass of unseen objects.

\begin{figure*}
 \includegraphics[width=1.00\textwidth]{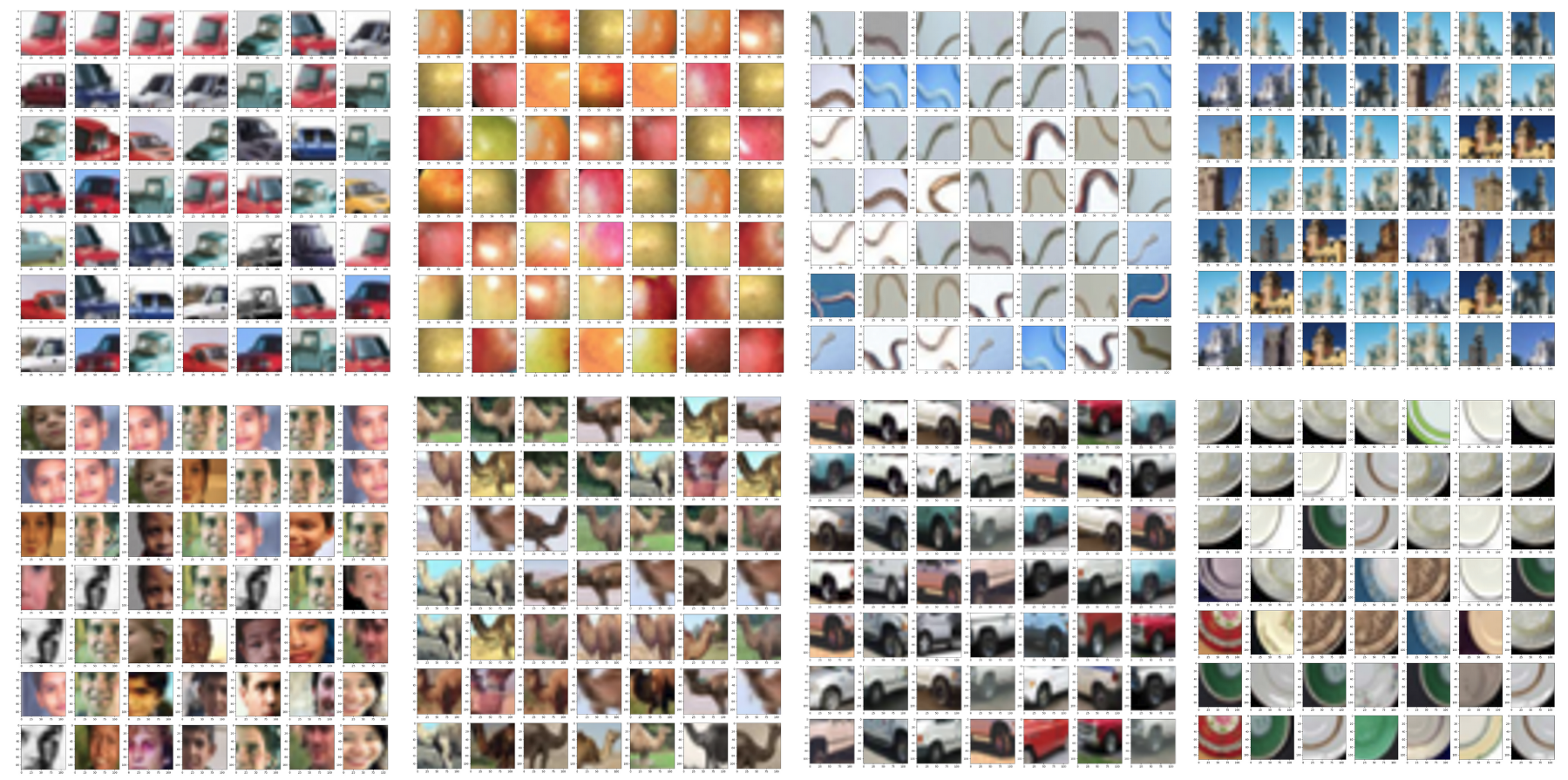}

 \caption{Visualization of our appearance + positional embedded clustering: Each block comprises patches that belong to an appearance cluster. CIFAR100 with 112x112 as patch size}
 \label{fig3:Features}

\end{figure*}

 \begin{figure*}
     \includegraphics[width=1.00\textwidth]{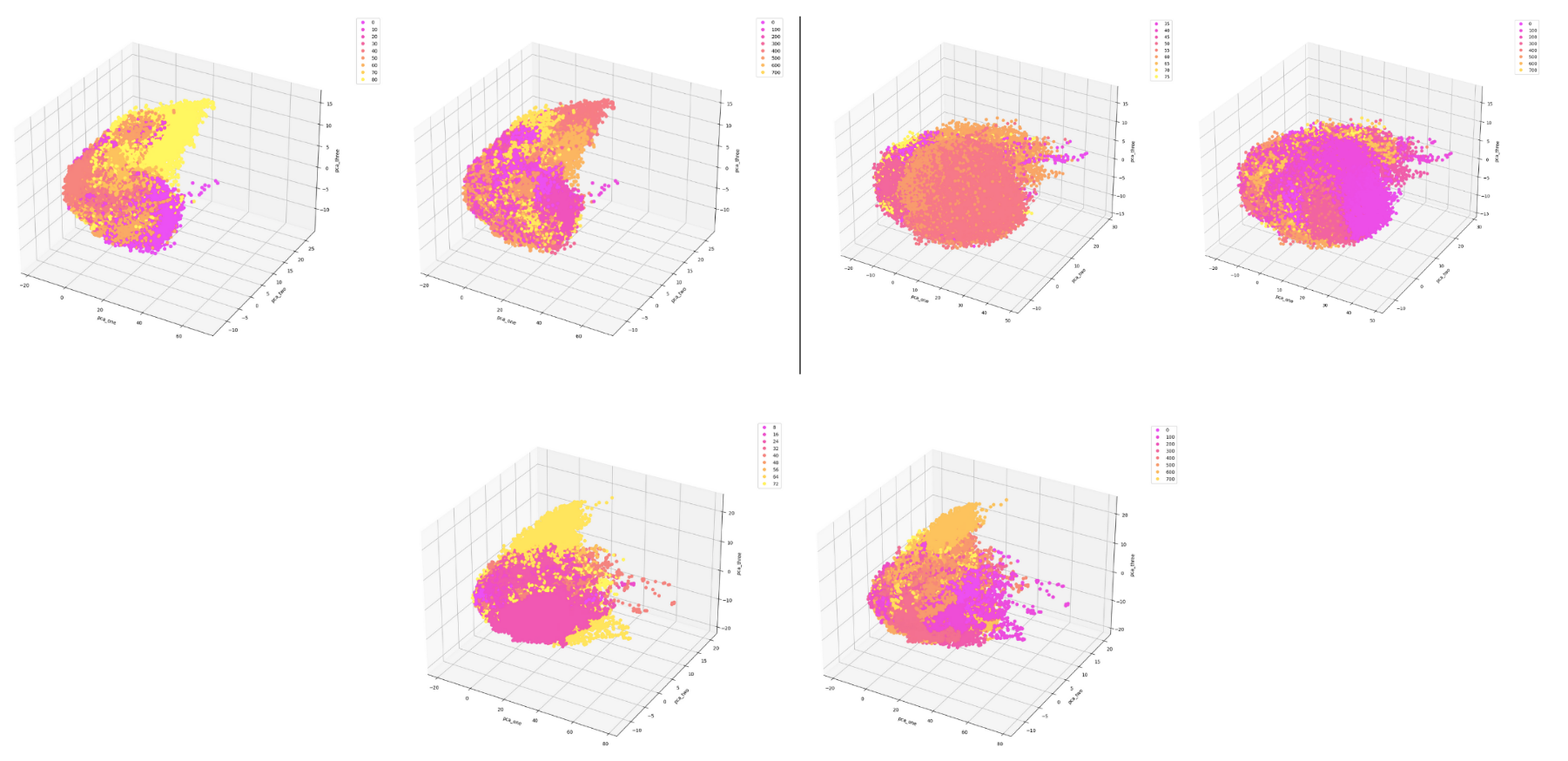}
     \caption{T-SNE feature plots for randomly generated data points from CIFAR100. {\it left}: clustered data points {\it vs.} {\it right}: semantic labels of the data points. Appearance vectors of each patch is represented as a data point. It is evident that given same number of cluster used, there is a significant levels of entropy in the grouping (right side) when using semantic labels as opposed to the appearance based clustering alone.  }
     \label{fig:tsne}
 \end{figure*}

\section{Model}

\subsection{Fix Notations}
Given an Image $I \in \mathcal{D}$, where $\mathcal{D}$ is our dataset of images, we obtain a subspace representation $F$ by feeding it to $f(x)$. In this paper, for $f(x)$ we use a vision transfomer~\cite{caron2021emerging}. The image $I$ is first divided into $\{M \times M\}$ or $M^2$ patches, and we extract features $F \in \mathbb{R}^{ M^2 \times N}$ from the patches, where $I \in \mathbb{R}^{224 \times 224 \times 3}$ and $N$ is the feature dimension. 

Each patch $P$ is represented as $\{P_m \in \mathbb{R}^{16 \times 16 \times 3} | _{m=1...M^2}\}$. In total there are $K$ clusters and $C$ cluster centers, where $\{C^k \in \mathbb{R}^{768}|_{k \in K}\}$. 

\subsection{Dataset Representation}

For all our open set recognition experiments, we assume that at test time we encounter novel object classes instead of instances from known object classes. For that purpose, we reorganize the datasets $\mathcal{D}$ into two non-overlapping subsets $\mathcal{D_K} = {(I_k, y_k, z_k)}^K_{k=1} \in \mathcal{X}, \mathcal{Y_K}$ and $\mathcal{D_U} = {(I_u, y_u, z_u)}^U_{u=1} \in \mathcal{X}, \mathcal{Y_U}$, where $\mathcal{D_K}$ and $\mathcal{D_U}$ are known and unknown class based data split and $\mathcal{Y_K} \cap \mathcal{Y_U}=\emptyset$.

$I$ \& $y$ are {\it images} and {\it class labels} respectively, and $z$ is the {\it superclass labels} which are obtained by creating a semantic {\it 2-tier} hierarchy of existing object classes. We used $z$ for whichever dataset it was already available (eg: {\it Cifar100}), and created or modified when such hierarchy was unavailable. Thus our method learns to reason object instances from $\mathcal{D_U}$ at test time, using information learned from known split $\mathcal{D_K}$ of the data, where $\mathcal{D_U}$ is untouched at learning.

\subsection{Appearance Based Grouping}\label{sec:grouping}
Extracted feature grid $F \in \mathbb{R}^{ M^2 \times N}$ for an image $I$, is then gathered by patches and clustered in a class agnostic manner, into {\it K} clusters. Each feature patch $f^{i,m} \in \mathbb{R}^N |_{f \in F}$ belongs to $i^{th}$ image ($i \in I$) and $m^{th}$ location (where $ m \in M^2$). Optionally, we also embed the location information of the patch to its feature vector, represented as $f^{i,m,l} \in \mathbb{R}^N |_{f \in F}$, where we compute a 2-dimensional positional encoding $\{sin(x), cos(y)\}$ using the technique proposed ~\cite{vaswani2017attention}, where $x$ \& $y$ are the 2D positions of the patch. $N$ is the feature dimension, and for all our experiments $N$ is set at 768. With the size of image $I$ at $224 \times 224$ and patch $P$'s size $16 \times 16$, we get $M = 14$; a total of $14 \times 14$ patches, (196 patches in total for an image $I$). 

We then cluster the gathered patch-wise features into $K$ clusters, where each cluster center is represented by $\{C^k \in \mathbb{R}^{768} | _{k \in K} \}$. Using the clusters, we then compute {\it semantic confidence vector} for each cluster by summing up the number of patches belonging to object classes and normalizing them as,

\begin{equation}\label{eq:one}
    S_k = \frac{1}{Q^k} \sum_{n=1}^{Q^k} \mathbb{G}(f_c) \quad where 
    \begin{cases} 
      \mathbb{G}(f_c) = 1 & if \; P_m \in G \\
      0 & otherwise 
   \end{cases}
 \quad \quad where  \quad  P \in \mathbb{R}^G \;, \; S  \in \mathbb{R}^{G \times K}
\end{equation}

where $Q^k$ is the number of patches belonging cluster $K$, and $G$ is the list of classes where $\mathbb{G}(f_c)$ is valid if the patch belong to the class $G$. Doing so allows us to represent each cluster as a histogram of class labels, and the softmax or normalization allows them to be used as a confidence vector. Using above equation~\ref{eq:one}, we construct $S \in \mathbb{R}^{G \times K}$ is our prior that we learn from the training set which we use at inference to reason or model unseen object classes.

\begin{table*}[tbhp]
\centering
\footnotesize
\begin{tabular}{c c c c c c c} 
 \hline
 & Clusters & Patch size & Top 1 (\%) ↑ & Top 2 (\%) ↑ & Top 3 (\%) ↑ \\ [0.5ex] 
 \hline\hline

CIFAR100 & 200 & 32 & 61.39 & 81.70 & 91.15 \\
CIFAR100 & 30 & 112 & 59.65 & 86.02 & 94.58 \\
CIFAR100 & 50 & 112 & 63.35 & 85.80 & 94.52 \\
CIFAR100 & 200 & 224 & 65.70 & 89.24 & 95.41 \\
CIFAR100 & 200 & 112 & 69.02 & 90.19 & 96.79 \\
CIFAR100 & 800 & 112 & 70.66 & 91.86 & 97.11 \\
CIFAR100 (saliency added) & 800 & 112 & 70.63 & 91.85 & 97.10 \\
CIFAR100 (with \cite{vaze2022gcd} features*) & 800 & 112 & 67.93 & 88.89 & 95.53 \\
ImageNet64x64 & 1000 & 224 & 77.50 & 90.95 & 95.62 \\
ImageNet64x64 & 1000 & 112 & 84.80 & 95.00 & 97.82 \\
ImageNet64x64 & 1500 & 112 & 84.23 & 94.69 & 97.67 \\

  \hline
\end{tabular}
\caption{Superclass estimation accuracy: K-means clustering on CIFAR100 and Imagenet dataset with various clusters and patch sizes. * is the result from clustering features obtained from \cite{vaze2022gcd}}
\label{table:1}
\end{table*}

\begin{table*}[tbhp]
\centering
\footnotesize
\begin{tabular}{c c c c c c c} 
 \hline
 & Clusters & Patch size & Patch Embedding Weights & Top 1 (\%) ↑ & Top 2 (\%) ↑ & Top 3 (\%) ↑ \\ [0.5ex] 
 \hline\hline

CIFAR100 & 800 & 112 & 0 & 70.66 & 91.86 & 97.11 \\
CIFAR100 & 800 & 112 & 0.3 & 71.95 & 91.87 & 97.6 \\
CIFAR100 & 800 & 112 & 0.5 & 70.43 & 91.63 & 97.54 \\
CIFAR100 & 800 & 112 & 0.7 & 69.83 & 91.36 & 97.23 \\
ImageNet64x64 & 1000 & 112 & 0 & 84.80 & 95.00 & 97.82 \\
ImageNet64x64 & 1000 & 112 & 0.3 & 85.02 & 95.24 & 97.97 \\
ImageNet64x64 & 1000 & 112 & 0.5 & 83.18 & 94.34 & 97.73 \\

  \hline
\end{tabular}
\caption{Superclass estimation accuracy: K-means clustering on CIFAR100 and Imagenet dataset with various clusters and patch sizes with positional embedding.}
\vspace{-2mm}
\label{table:2}
\end{table*}

\begin{figure*}
 \centering
 \includegraphics[scale=0.45]{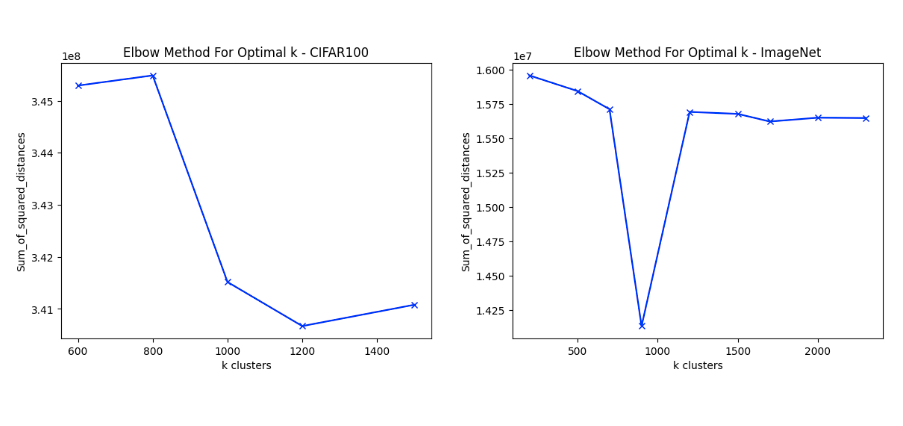}
 \vspace{-10mm}
 \caption{Optimal K via elbow method for Cifar100 \& imagenet datasets}
 \label{fig:elbow}

\end{figure*}

\subsection{Inference}

At query time, when encountered a test image $I^t \in \mathcal{D_U}$ containing an object that is unknown at train time, we extract features $F^t \in \mathbb{R}^{ M^2 \times N}$ using our vision transformer $f(x)$. The distances between features and cluster centers $C$ are then computed using,

\begin{equation}\label{eq:dist}
    D_k^m = argmin_{k}|| f^{i,m,l} - C^k ||_2
\end{equation}

The final {\it semantic vector predictions} $\mathbb{P}$ are obtained using,
\begin{equation}
    \mathbb{P}_{I^t} = \frac{1}{M^2}\sum_{m=0}^{M^2} S(D_k^m)
\end{equation}

where we find the closest cluster center $C^k$ for each of the patch using the features of the patch $f^{i,m,l}$. The semantic confidence vectors $S  \in \mathbb{R}^{G \times K}$ computed using~\ref{eq:one}, essentially represents how the appearances of each cluster is distributed across semantic class labels. At test time given~\ref{eq:dist} computed for the test image,  we average across all the {\it semantic confidence vectors} for all patch in the test image (whose object class is unseen), to obtain a final semantic prediction $\mathbb{P}_{I^t}$. The semantic prediction vector $\mathbb{P}$ essentially quantifies similarities between the unseen object class of the test instance and all the known classes, taking into account both appearance and 2D positional information. We then group the semantic positional vector by superclasses (using superclass hierarchy created) to get the superclass prediction. For example, lets assume a test instance to contain a {\it forklift}, which can have a semantic positional vector like \{{\it car: 0.2, truck: 0.3, bike: 0.05,..., bird: 0.0}\}, and the subsequent superclass prediction can be \{{\it vehicles: 0.7, furnitures:0.1, animals: 0.05, birds:0.0... }\}.


\section{Datasets}
For our experiments, we primarily use {\it CIFAR100} \& {\it Imagenet}~\cite{russakovsky2015imagenet} datasets, two of the most commonly used datasets. The choice of the datasets were primarily based on availability of diverse and large number of object classes within, and on ease of reproducibility of results. Both {\it CIFAR100} \& {\it Imagenet} contain over 100 object classes each, including some of rare objects, making it an ideal choice for our experiments.

In CIFAR100, 100 object classes of CIFAR100 were grouped into 6 superclasses namely, vehicles, vegetation, land animals, water animals, structures, each supercluster containing 12-18 non overlapping classes each. Similarly, {\it Imagenet} contains 1000 classes overall, but for our experiments, we use an available subset of the ImageNet 64x64 with 168 object classes. Please see the supplementary material for detailed class hierarchy information, which will be also released upon the acceptance of the paper.

 \begin{figure*}[hbt!]
 \centering
 \includegraphics[scale=0.35]{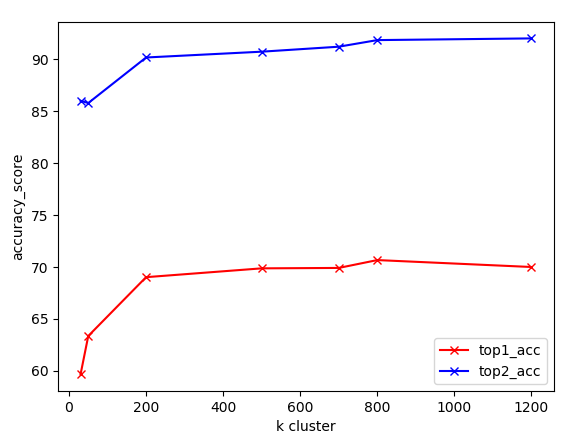}
 \vspace{-3mm}
 \caption{CIFAR100: Number of clusters vs {\it top-1} and {\it top-2} accuracies}
 \label{fig:clusters_v_acc}

 \end{figure*}

\begin{figure*}
 \centering
 \includegraphics[scale=0.35]{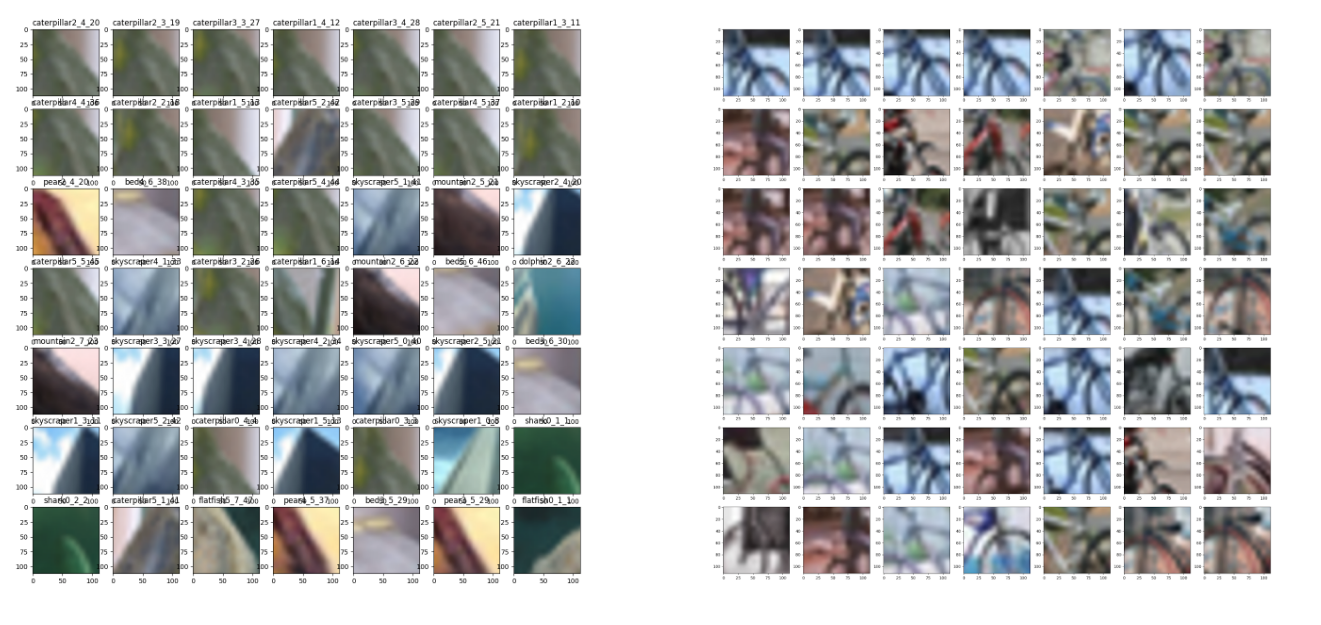}
 \vspace{-7mm}
 \caption{Qualitative comparison between low-level (left) \& mid-level appearance cues (right).}
 \label{fig:low_v_mid}

\end{figure*}

\section{Implementation Details}

For feature extraction, we primarily use ViT Dino~\cite{caron2021emerging}. Images are resized to a standard size of 224 x 224, along with mean shift \& center crop transformations applied to mimic the ViT Dino transformations. These transformed images are fed into ViT Dino, which chunks the image further into 16 x 16 pixel patches, to a total of 196 non-overlapping patches. The chunked patches are position-encoded and fed into ViT Dino~\cite{caron2021emerging}, which outputs a 197 $\times$ 768-dimensional vector. The output appearances descriptors are 768-dimensional vectors for each patch plus a additional global feature vector of the same dimension. For our experiments, we do not use the $1^{st}$ feature vector which embeds global feature cues, and instead use the rest 196 vectors, for which we have a known 2D positional association. In other words, we only use the patch-wise appearance vectors for which we have the direct mapping in the image space. Given that we have 196 appearance vector for each image, we gather all the appearance vectors of all images across the dataset, and cluster using iterative K-means clustering with mini-batches of size 6000 images. 

Positional embedding detailed in section~\ref{sec:grouping} is done using the technique detailed in~\cite{vaswani2017attention}. For obtaining optimal $K$ for K-means, we find the sum of squared distance error for a range of K clusters and verify the region by picking the lowest and comparing with other range points. The accuracies we obtained for various $K$ clusters reflect the K value estimated via elbow method shown in Figure~\ref{fig:elbow}. 

We also trained a supervised contrastive network using supervision loss for class label regression introduced by~\cite{vaze2022gcd}, for which we used the default configurations used by the ViT Dino~\cite{caron2021emerging}, and subsequent feature extraction \& clustering follows the aforementioned steps for fair comparison (results reported in~\ref{table:1}. Also, we have implemented a saliency metric to prioritize more salient regions {\it vs.} background regions. For that, we computed a channel-wise sum of the feature vectors and normalize to obtain a 2D weight vector (with size same as the 2D feature vector) which is then multiplied with the semantic confidence vectors (results reported in~\ref{table:1}.

\section{Results and Discussion}
Figure~\ref{fig:tsne} highlights the discrepancy between grouping based on the semantic labels {vs.} visual description based grouping. It is evident that the representation based on semantic grouping is not compact. This further strengthens our claim that visual features space does not always overlap with the semantic space. Our experiments aim to deconstruct the visual appearance space into low \& mid-level features, and aims to learn a novel object representation that groups appearance cues, which in turn allows us to reason any novel or unseen object as a combination of known mid-level appearance cues. Also Figure~\ref{fig3:Features} shows qualitative representation of patches clustered together. These patches are mid-level appearance cues that are somewhat generic and building a comprehensive set of such appearance cues would allow us to model any novel object.

Figure~\ref{fig:elbow} shows how the number of clusters are estimated. We employed the elbow method that maximizes the compactness of the clusters. As we see from the Figure~\ref{fig:elbow}, the number of optimal clusters are almost similar for both CIFAR100 (left) and Imagenet datasets (right). Despite Imagenet having {\it 3x} more images and 50\% more object categories, the optimal K estimated is not considerably different between the two. This further bolsters our claim that there are only a finite number of mid-level visual cues and most objects can be reliably represented using those, including novel or unseen objects. It is also evident from Figure~\ref{fig:clusters_v_acc} that agrees with the Figure ~\ref{fig:elbow}, where the accuracy saturates after the optimal cluster sizes is reached. This clearly shows that using a set of general-purpose mid-level appearance cues is sufficient learn any object representation.

Table~\ref{table:1} shows the {\it Top 1,2,3} accuracies for both the datasets with varying patch sizes and number of clusters without using positional embedding, whereas Table~\ref{table:2} shows the top-N accuracies with positional embeddings added with varying weights, using~\cite{vaswani2017attention}. We observe the number of clusters estimated by the elbow method based on compactness yields highest accuracies for both datasets. We observed that the smaller patches (32 x 32) are comprised of less descriptive features as opposed to larger patches with more descriptive mid-level cues. The qualitative comparison between the low \& mid level cues is shown in Figure~\ref{fig:low_v_mid}. Using patch size of 112 yields the highest accuracy for both datasets with and without positional embedding. In Table~\ref{table:2} we show the use of positional embedding results in considerable improvement in the accuracies, showing the significance of using the location cues along with appearance cues. 

It also has to be noted that, we train an end-to-end transformer by adding supervised contrastive loss (referred with * in Table~\ref{table:1}) proposed by~\cite{vaze2022gcd} which performs inferior to the pretrained ViT Dino. This shows that fine-tuning the networks to encode additional supervision (of known classes) is overfitting the network to the set of known object classes and is affecting the generalizability of the network. In contrast, other data in Table~\ref{table:1} show that ViT Dino trained on contrastive loss alone learns to capture generic features that more reliably represents the visual world for reasoning novel objects. In addition, we observed adding saliency factor into our feature based clustering did not improve the results significantly. This could be because our saliency computation is simply the gradient of activation and is not as sophisticated as GradCAM~\cite{selvaraju2017grad} which requires class information. 

\section{Conclusion \& Future Work}

In this paper, we proposed a novel way of representing objects by dissecting them into mid-level appearance cues, and leveraging the appearances learned from known object classes to reason unknown object classes. The evaluation clearly show how conventional closed set assumptions fail to generalize for real-world use cases, and how the proposed approach for reasoning objects can effectively model any number of real world objects without need for large datasets that supervised methods do. 

As future work, we aim to build end-to-end trainable model that combines grouping within the transformer architecture using differentiable techniques for clustering. We also aim to build a complete object detection network using the proposed classifier as a second-stage of a two-stage network and demonstrate its applications on real-world problems.

{
\bibliographystyle{ieee_fullname}
\bibliography{egbib}
}

\end{document}


\title{Unseen Object Reasoning with Shared Appearance Cues}  

\section{Ablation Studies}
\label{ablation_study}

\subsection{Various K clusters}
Apart from the elbow-method to find optimal K clusters, we ran the experiment with different K means clusters and recorded the accuracies for both CIFAR100 Table\ref{table:2} and Imagenet Table\ref{table:one} . We also performed qualitative evaluations through visualizations of unknown set (evaluation data) belonging to respective superclass groups as shown in Fig.\ref{fig:eval_cluster} where each superclass is a combination of known classes (also explained under section \ref{alg:pseudo})

We notice that for both the datasets, CIFAR100 (with 80 training classes) and ImageNet64x64 (with 126 training classes), we hit a saturation in terms of accuracies with clusters approx around 800-1000, irrespective of the size of the dataset.

\subsection{Saliency map}
Saliency grid or saliency map helps in defining the salient regions to look for and weight higher for clustering. We performed the study by counting each patch with its normalized saliency weight wrt the image it belongs to, while assigning class distributions to the clusters. This way we made sure to not disturb the mean of the distribution but still weight patches depending on their saliency.

We notice weighing based on saliency does not affect the results, i.e., it neither improves nor deteriorates our results obtained otherwise for both CIFAR100 and ImageNet64x64 dataset.

We also show results for using saliency map for CIFAR100 unknown split images during evaluation in our table\ref{table:2} with a slight decrease in performance.

\subsection{Positional Embedding}
Each feature patch we used for clustering belong to a region in image space. We define positional embeddings for each patch based on the row and column it belongs to. Using weighted positional embedding as explained in the paper along with the features does show a pattern where for both CIFAR100 and ImageNet dataset, we see that the accuracies increase when we consider a small fraction (0.3 in our cases) of positional embeddings. However, if we increase the weight further, we notice constant decrease in performance. Our explanation is that all images in a dataset follow certain object distribution all along in a manner that top left of each image would give us some unique representation as compared to center or bottom right. These locations aid our clustering algorithm for CIFAR100 and ImageNet.

\subsection{Supervised contrastive clustering}
To evaluate the performance of \cite{vaze2022gcd} supervised contrastive loss using our clustering method, we obtain image features from the method proposed in \cite{vaze2022gcd} paper. We notice a decrease in performance with the features obtained from supervised contrastive loss when compared with features directly obtained from \cite{caron2021emerging}.

\subsection{Normalizing across entire dataset}
After finding K clusters, we defined each cluster as a combination of known classes based on number of patches that belong to it. For each cluster we normalised its distribution separately. However, we also ran experiment to evaluate normalizing cluster distribution across entire dataset vs eachcluster separately and we notice a slight increase in performance for CIFAR100, but for ImageNet there wasn't any improvement noticed.

\begin{table*}[hbt!]
\centering
\footnotesize
\begin{tabular}{c c c c c c c} 
 \hline
 & Clusters & Patch size & Top 1 (\%) ↑ & Top 2 (\%) ↑ & Top 3 (\%) ↑ \\ [0.5ex] 
 \hline\hline

ImageNet64x64 & 500 & 112 & 83.44 & 94.38 & 97.52 \\
ImageNet64x64 & 950 & 112 & 84.89 & 95.05 & 97.79 \\
ImageNet64x64 & 1000 & 224 & 77.50 & 90.95 & 95.62 \\
ImageNet64x64 & 1000 & 112 & 84.80 & 95.00 & 97.82 \\
ImageNet64x64 & 1500 & 112 & 84.23 & 94.69 & 97.67 \\
ImageNet64x64 & 2000 & 112 & 84.73 & 95.26 & 98.00 \\
ImageNet64x64 & 2500 & 112 & 85.67 & 95.46 & 98.04 \\

  \hline
\end{tabular}

\caption{ImageNet 64x64; Top 1, top2 and top 3 scores for different K clusters}
\label{table:one}
\end{table*}

\begin{table*}[tbhp]
\centering
\footnotesize
\begin{tabular}{c c c c c c c} 
 \hline
 & Clusters & Patch size & Top 1 (\%) ↑ & Top 2 (\%) ↑ & Top 3 (\%) ↑ \\ [0.5ex] 
 \hline\hline

CIFAR100 & 30 & 112 & 59.65 & 86.02 & 94.58 \\
CIFAR100 & 50 & 112 & 63.36 & 85.81 & 94.53 \\
CIFAR100 & 200 & 224 & 65.70 & 89.24 & 95.40 \\
CIFAR100 & 200 & 112 & 69.03 & 90.19 & 96.79 \\
CIFAR100 & 200 & 32 & 61.39 & 81.70 & 91.15 \\
CIFAR100 & 500 & 80 & 68.35 & 89.47 & 96.86 \\
CIFAR100 & 500 & 112 & 69.875 & 90.75 & 97.13 \\
CIFAR100 & 500 & 144 & 68.56 & 91.67 & 97.04 \\
CIFAR100 & 500 & 224 & 65.50 & 87.50 & 93.98 \\
CIFAR100 & 700 & 112 & 69.91 & 91.23 & 97.09 \\
CIFAR100 & 800 & 64 & 70.20 & 89.88 & 96.83 \\
CIFAR100 & 800 & 112 & 70.55 & 91.61 & 97.10 \\
\textbf{CIFAR10} & \textbf{800 (normalized across entire dataset)} & \textbf{112} & \textbf{70.667} & \textbf{91.87} & \textbf{97.10} \\
CIFAR100 & 800 (saliency map) & 112  & 70.53 & 91.60 & 97.10 \\
CIFAR100 & 800 (saliency map with eval*) & 112 & 70.58 & 91.62 & 97.11 \\
CIFAR100 & 800 (salient eval* + normalized across entire dataset) & 112 & 70.63 & 91.87 & 97.10 \\
CIFAR100 & 800 (positional embedding weight 0.3) & 112 & 71.95 & 91.87 & 97.10 \\
CIFAR100 & 800 (positional embedding weight 0.5) & 112 & 70.43 & 91.63 & 97.54 \\
CIFAR100 & 800 (positional embedding weight 0.7) & 112 & 69.83 & 91.36 & 97.23 \\
CIFAR100 & 800 (supervised contrastive loss from \cite{vaze2022gcd}) & 112 & 67.93 & 88.89 & 95.53 \\
CIFAR100 & 1200 & 112 & 70.00 & 92.03 & 97.58 \\
CIFAR100 & 1500 & 112 & 70.41 & 91.61 & 97.52 \\
  \hline
\end{tabular}

\caption{CIFAR100; Top 1, top2 and top 3 scores for different K clusters; explained under ablations study section \ref{ablation_study}. * saliency map used during evaluation as well. Best accuracy is shown in bold.}
\label{table:2}
\end{table*}

\begin{figure}
     \centering
     \begin{subfigure}[b]{0.5\textwidth}
         \centering
         \includegraphics[width=\textwidth]{cifar100_acc_top12.png}
         \caption{CIFAR100}
         \label{fig:cifaracc}
     \end{subfigure}
     \hfill
     \begin{subfigure}[b]{0.55\textwidth}
         \centering
         \includegraphics[width=\textwidth]{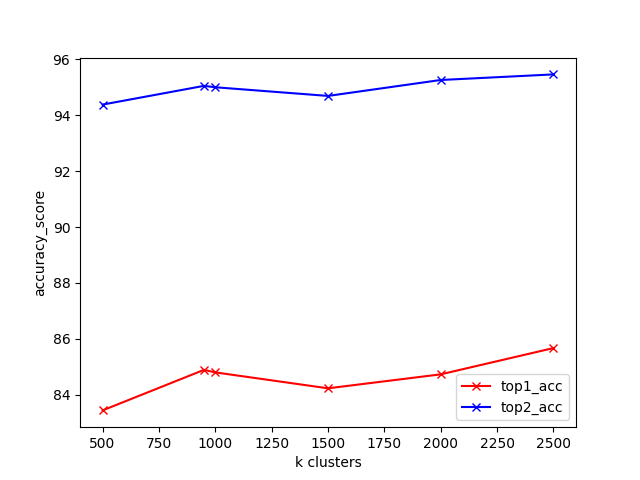}
         \caption{Imagenet 64x64}
         \label{fig:imagenetacc}
     \end{subfigure}
        \caption{Number of clusters vs {\it top-1} and {\it top-2} accuracies}
        \label{fig:two graphs}
\end{figure}

\begin{figure*}
 \centering
 \includegraphics[width=0.78\textwidth, height=0.78\textheight]{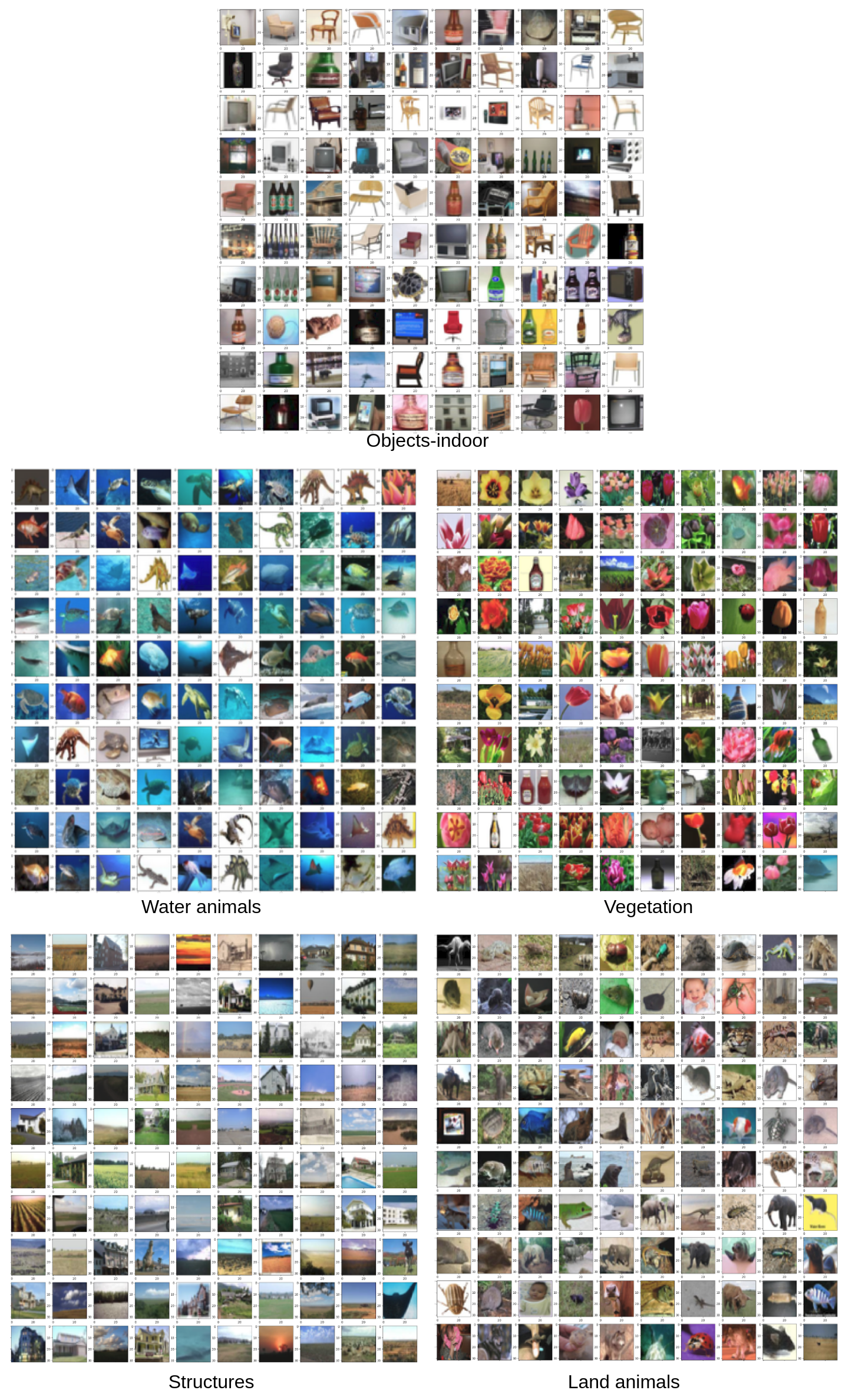}
 \vspace{-3mm}
 \caption{Qualitative Results of our superclass-classification model: Superclass clustering of CIFAR100 evaluation set (Unknown 20 classes split) based on the clustering learnt from the known 80 classes split. }
 \label{fig:eval_cluster}
 \end{figure*}


 \begin{figure*}
 \centering
 \includegraphics[width=0.75\textwidth, height=0.75\textheight]{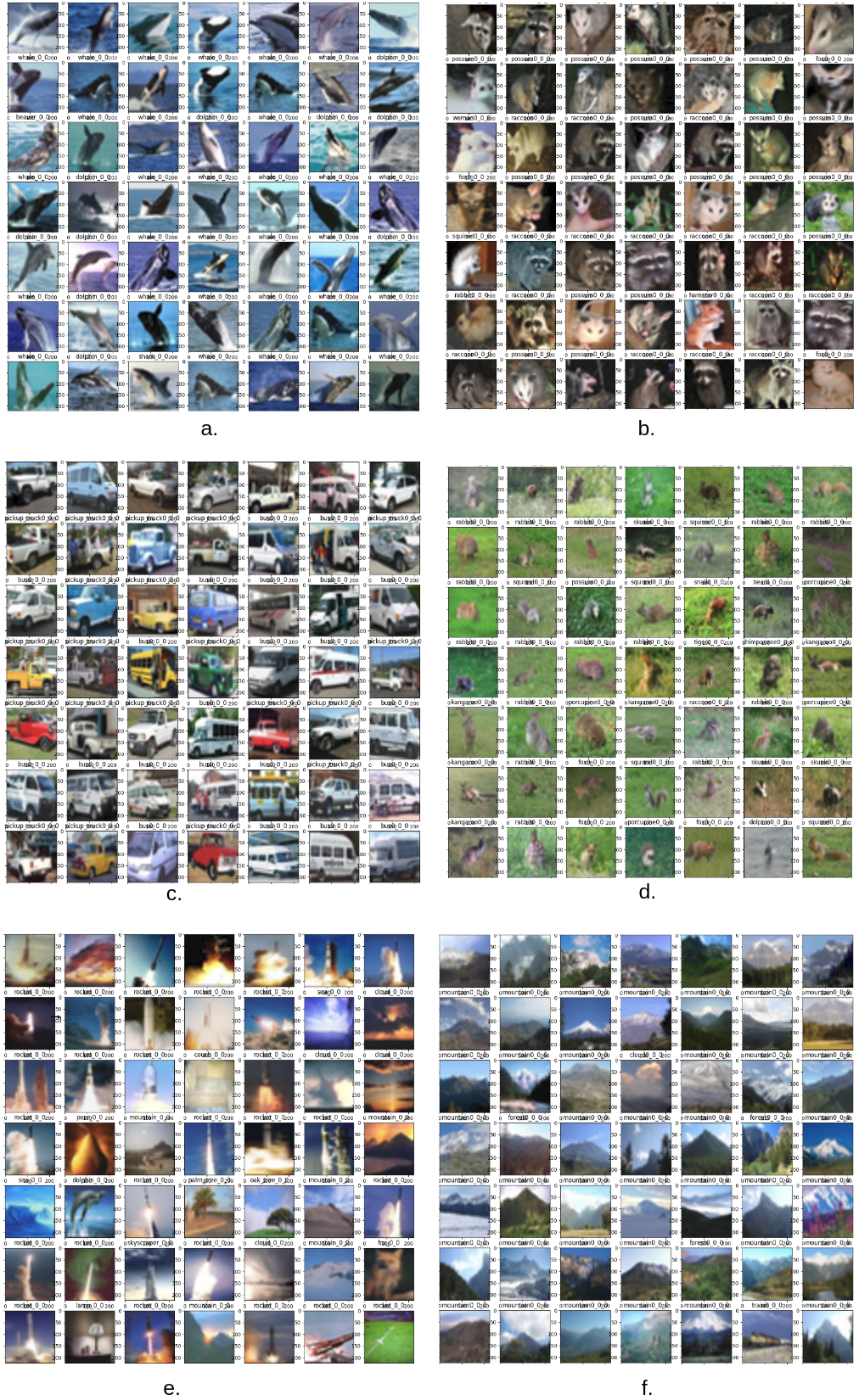}
 \vspace{-3mm}
 \caption{Known Data K-means clusters: CIFAR100 224x224 patches. We notice that clusterings take into account feature spaces which are similar. Observations noted - (b) a women in white dress, row 3 col 1, appears similar to other members in the cluster. (e) lamp with background light, row 7 col 2, and fruit pear against light, row 4 col 2 appear similar to other images in the cluster. (d), different classes of rodents have been captured, all with similar grassy background have been grouped.}
 \label{fig:clusters_v_acc_imgnet}
 \end{figure*}

\section{Pseudo Code}\label{alg:pseudo}
Psuedocodes for learning and inference are detailed respectively.
\begin{algorithm}
\caption{Learning Visual Space Reasoning of Object Appearances}\label{alg:cap}
\begin{algorithmic}
\Require Given a dataset of Images $I \in \mathcal{D_K}$ and labels $G$
    \For{each $I$ in $\mathcal{D_K}$}        
        \State Extract features $F^I \in \mathbb{R}^{M^2 \times N}$
        \State $F^I \gets \text{positional\_embedding}(F^I)$
        \State $F \gets F ; F^I$
    \EndFor
    \State $C^k = \text{KMeans}(F, K)$, where $\{C^k \in \mathbb{R}^{768} \mid k \in K\}$
    \State $S_k = \text{semantic\_confidence\_vector}(C^k, F^I, G)$ using Equation(1) $\forall \: k \in K$
\end{algorithmic}
\end{algorithm}

\begin{algorithm}
\caption{Inference}\label{alg:inf}
\begin{algorithmic}
\Require Given a dataset of unknown Images $I_t \in \mathcal{D_U}$, Learned Semantic Confidence Vectors $S_k$, and Cluster Centers $C$
    \For{each $I_t$ in $\mathcal{D_U}$}
        \State Extract features $F^{I_t} \in \mathbb{R}^{M^2 \times N}$
        \State $F^{I_t} \gets \text{positional\_embedding}(F^{I_t})$
        \State Distances $D_k^m = \arg\min_{k} \| F^{I_t} - C^k \|_2$
        \State Semantic Vector Predictions
        \State \quad $\mathbb{P}_{I_t} = \frac{1}{M^2} \sum_{m=0}^{M^2} S(D_k^m)$
    \EndFor
\end{algorithmic}
\end{algorithm}

\section{Visualization of Appearance clusters - tSNE figures}

We have generated additional {\it tSNE} figures (Figures 3 to 19) to visually demonstrate how much entropy exists between the appearance based clustering of features {\it vs.} semantic labels associated with them. These figures are random samples of data points color-mapped to corresponding clusters (left) and to the respective class labels (right). The number of classes and number of clusters are kept the same for fair comparison.  It is evident from the figures that appearance based grouping is significantly smoother as expected whereas the corresponding semantic labels-based grouping contains too much variance. As highlighted in the paper, this is primarily due to the similarities and dissimilarities that exist across and within semantic classes respectively.

 \begin{figure*}
     \includegraphics[width=1.00\textwidth]{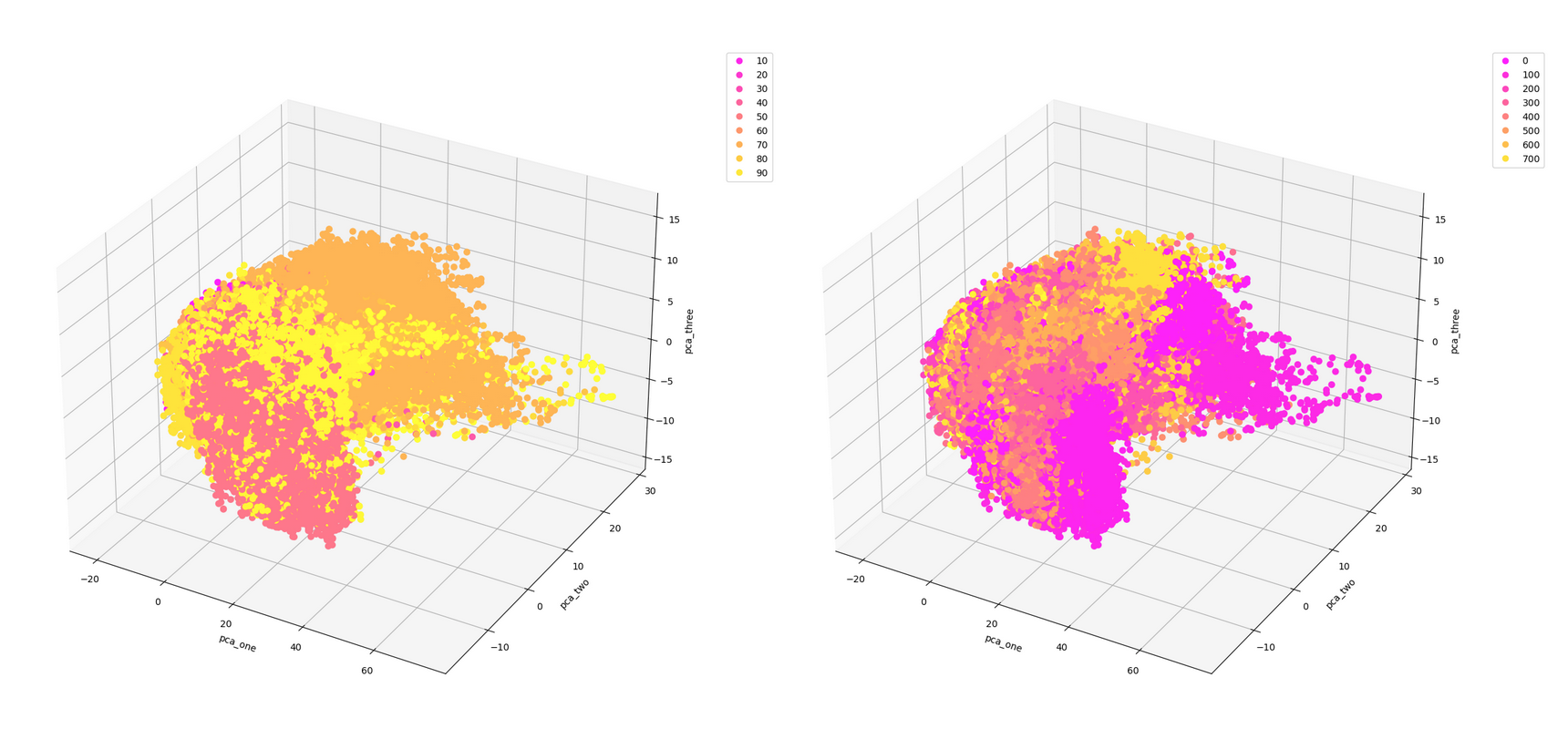}
     \caption{Additional T-SNE feature plots for randomly generated data points using same number of classes and appearance clusters, from CIFAR100. {\it left}: clustered data points {\it vs.} {\it right}: semantic labels of the data points. Appearance vectors of each patch is represented as a data point. It is evident that given same number of cluster used, there is a significant levels of entropy in the grouping (right side) when using semantic labels as opposed to the appearance based clustering alone.  }
     \label{fig:tsne1}
 \end{figure*}
  \begin{figure}
     \includegraphics[width=.50\textwidth]{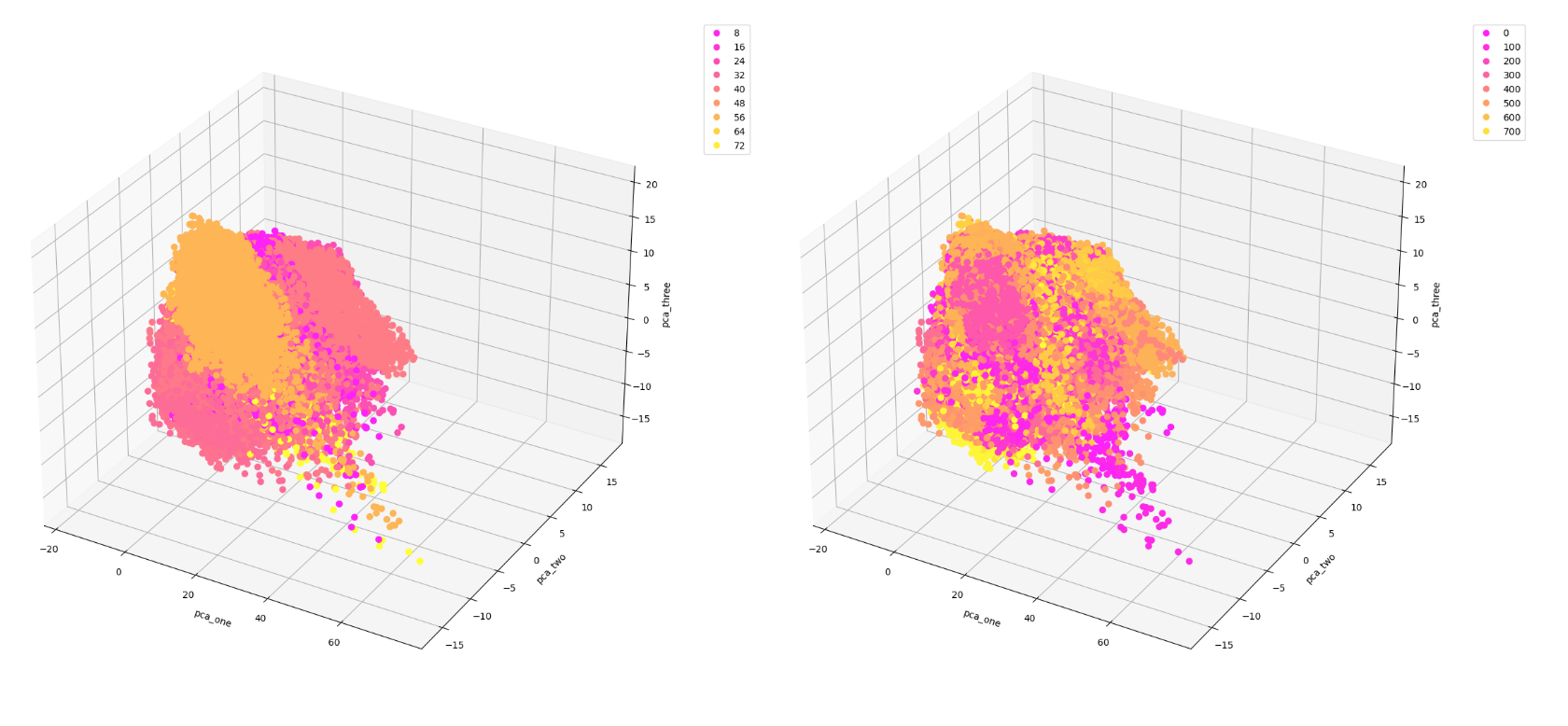}
     \caption{T-SNE feature plots for randomly generated data points using same number of classes and appearance clusters, from CIFAR100. }
     \label{fig:tsne2}
 \end{figure}
   \begin{figure}
     \includegraphics[width=.50\textwidth]{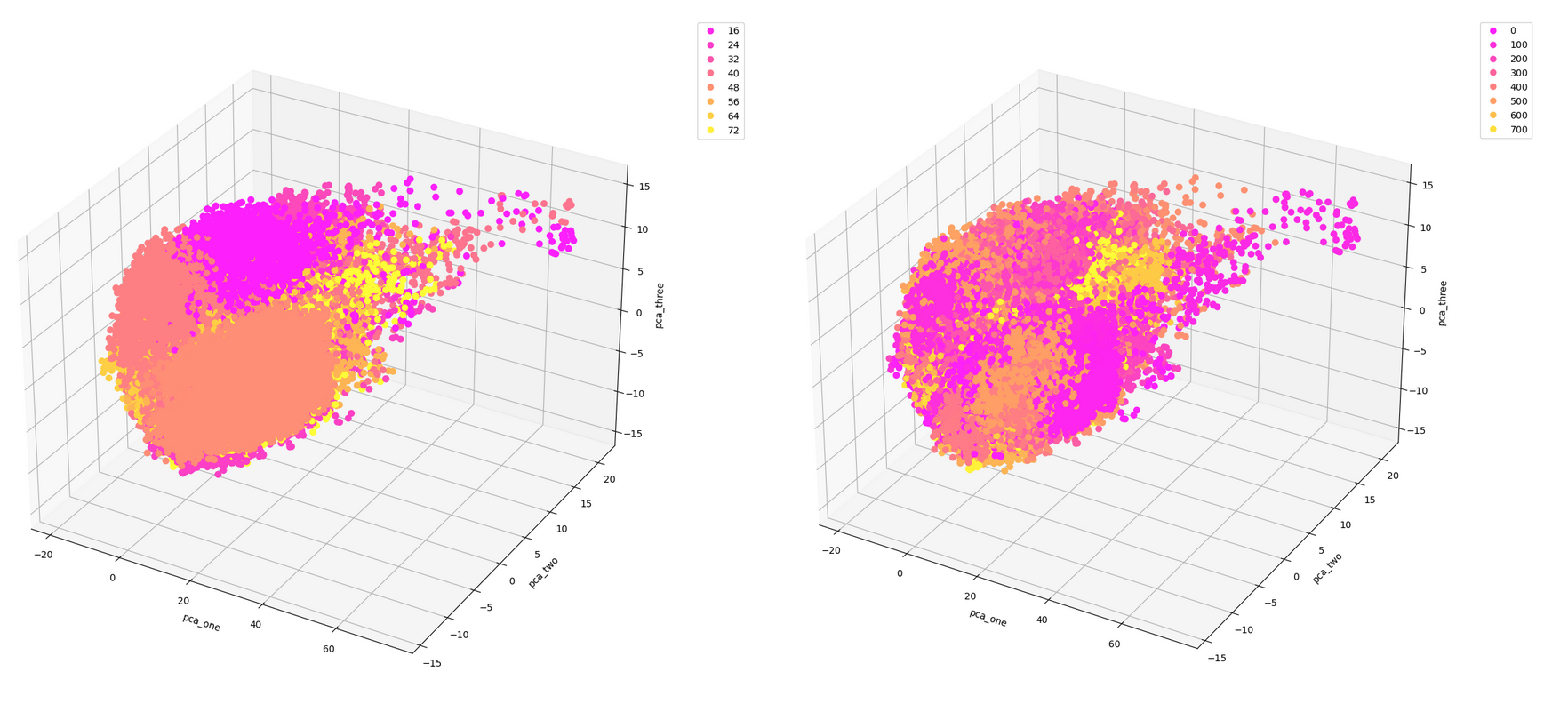}
     \caption{T-SNE feature plots for randomly generated data points using same number of classes and appearance clusters, from CIFAR100. }
     \label{fig:tsne3}
 \end{figure}
   \begin{figure}
     \includegraphics[width=.50\textwidth]{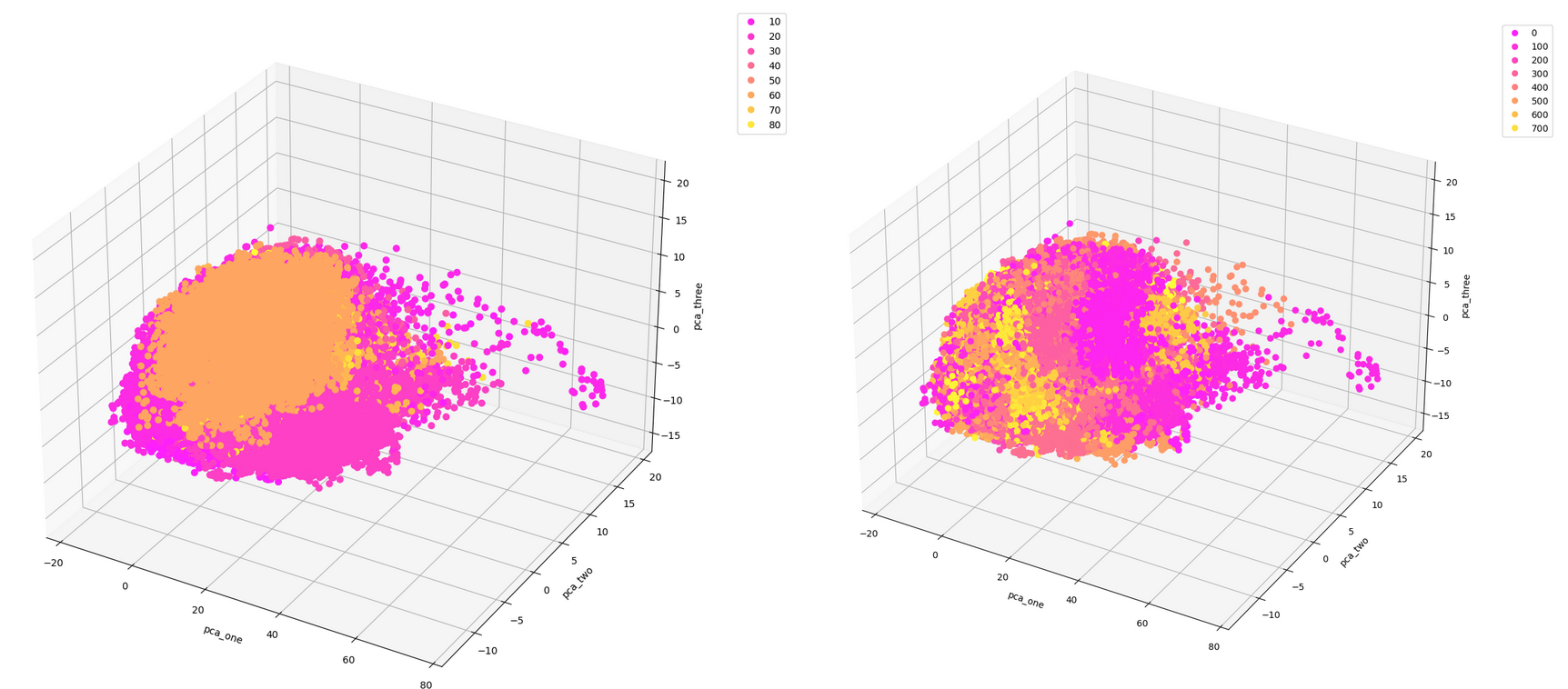}
     \caption{T-SNE feature plots for randomly generated data points using same number of classes and appearance clusters, from CIFAR100. }
     \label{fig:tsne4}
 \end{figure}
   \begin{figure}
     \includegraphics[width=.50\textwidth]{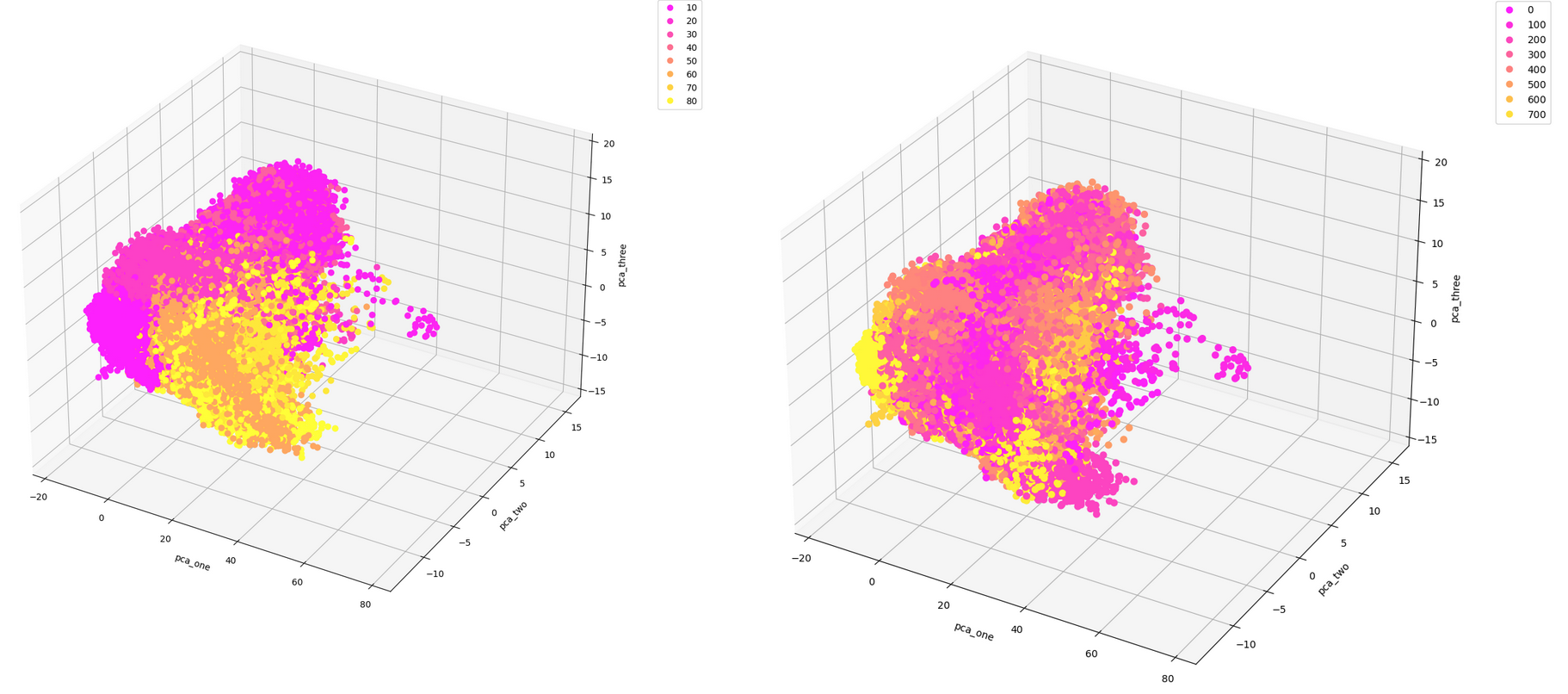}
     \caption{T-SNE feature plots for randomly generated data points using same number of classes and appearance clusters, from CIFAR100. }
     \label{fig:tsne5}
 \end{figure}
   \begin{figure}
     \includegraphics[width=.50\textwidth]{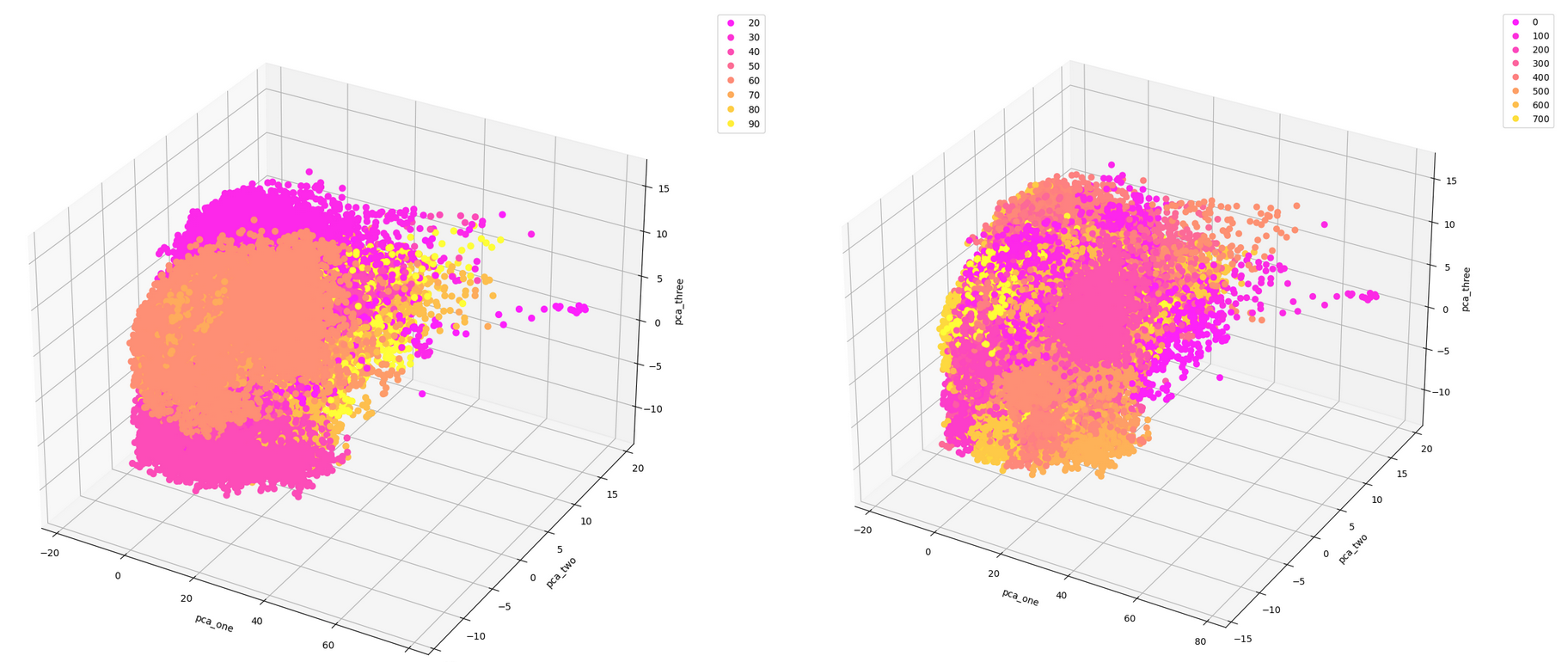}
     \caption{T-SNE feature plots for randomly generated data points using same number of classes and appearance clusters, from CIFAR100. }
     \label{fig:tsne6}
 \end{figure}
   \begin{figure}
     \includegraphics[width=.50\textwidth]{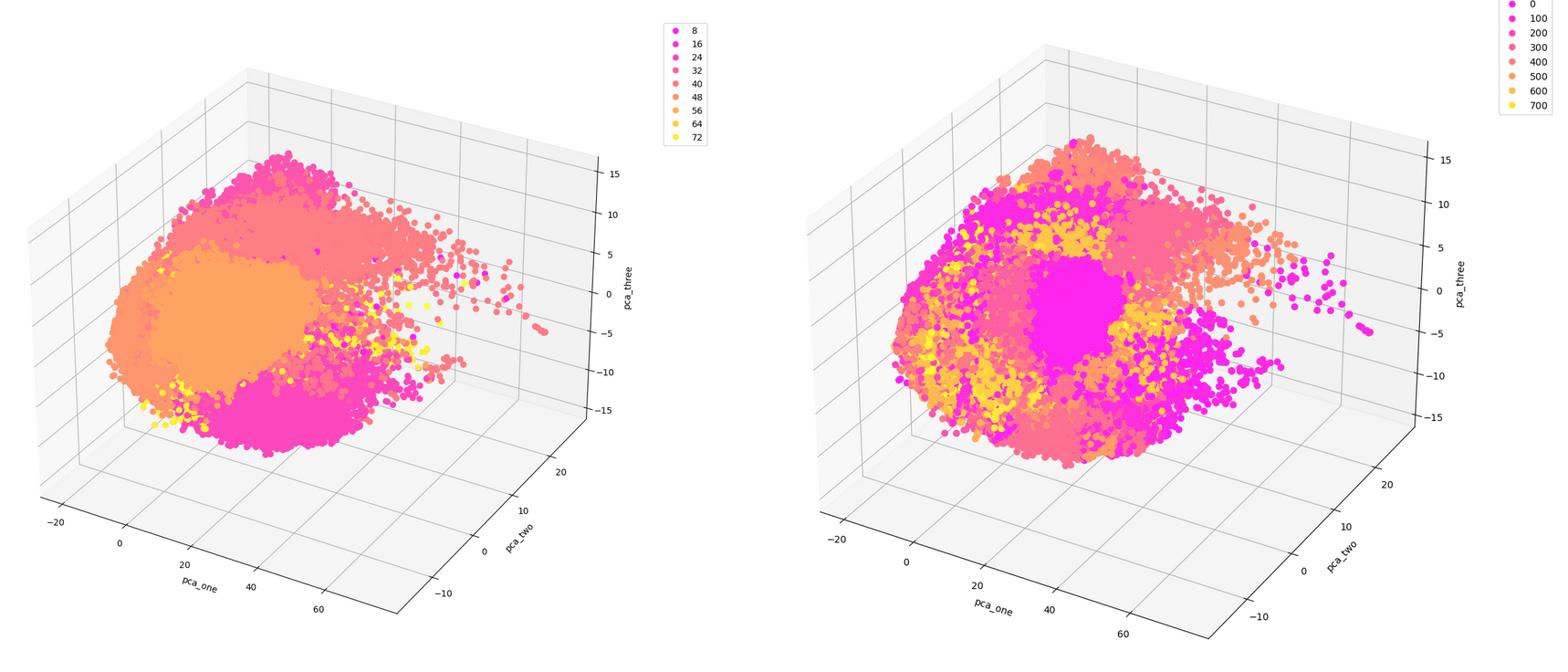}
     \caption{T-SNE feature plots for randomly generated data points using same number of classes and appearance clusters, from CIFAR100. }
     \label{fig:tsne7}
 \end{figure}
   \begin{figure}
     \includegraphics[width=.50\textwidth]{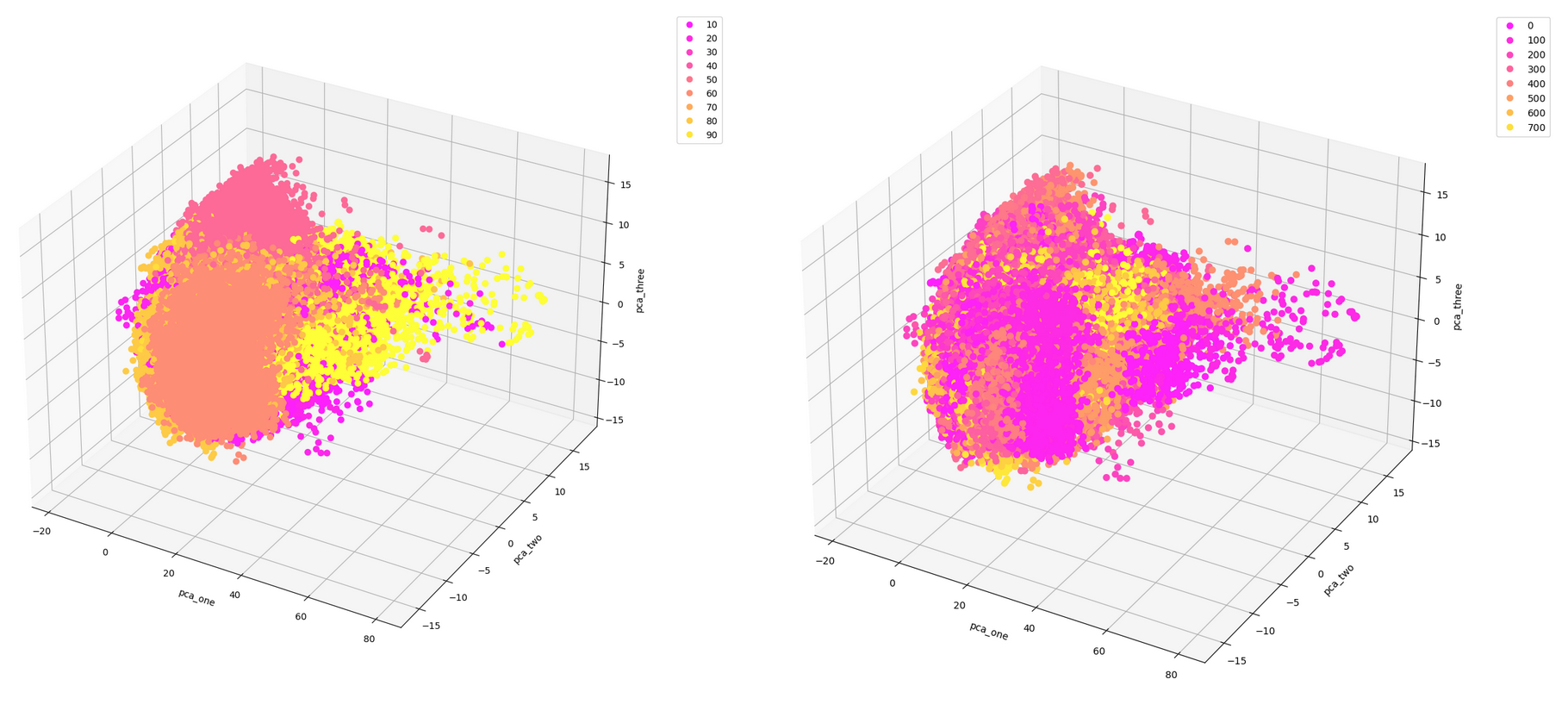}
     \caption{T-SNE feature plots for randomly generated data points using same number of classes and appearance clusters, from CIFAR100. }
     \label{fig:tsne8}
 \end{figure}
   \begin{figure}
     \includegraphics[width=.50\textwidth]{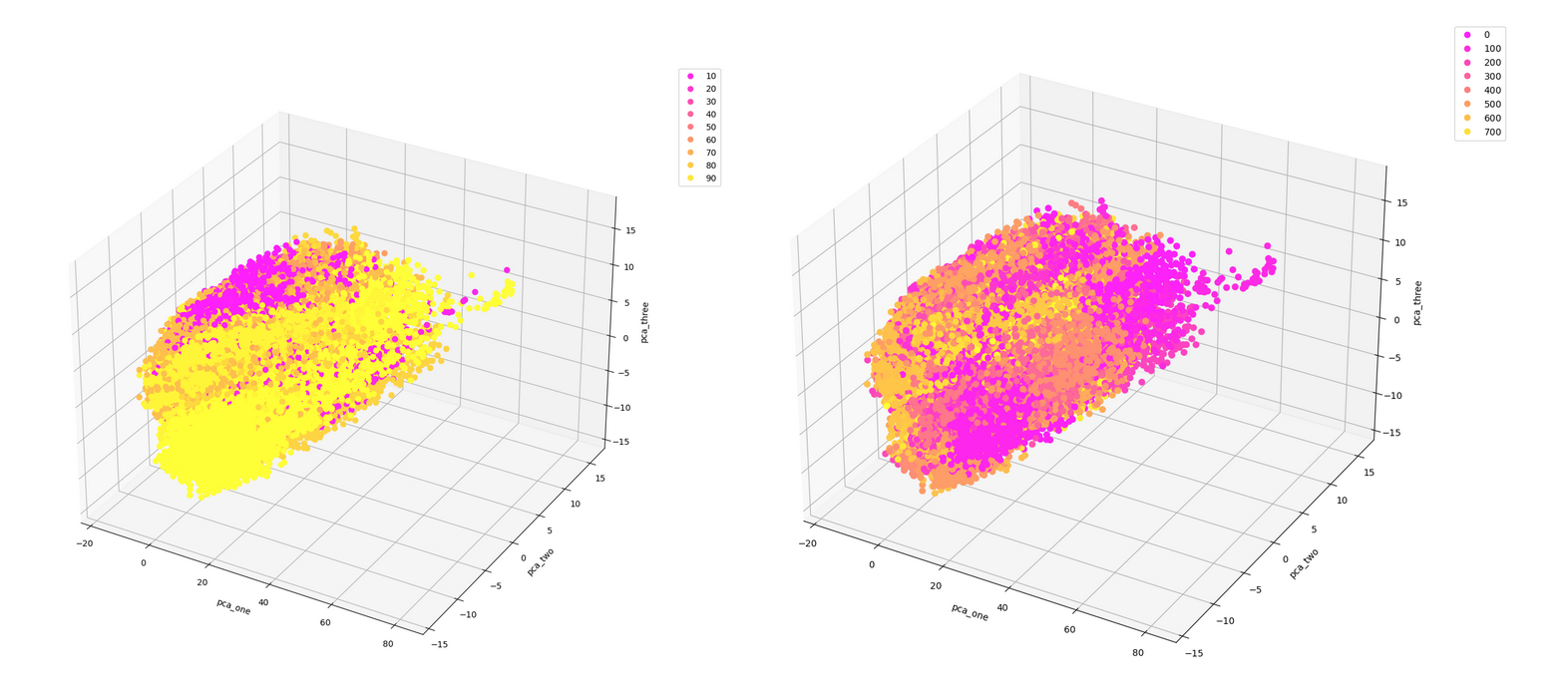}
     \caption{T-SNE feature plots for randomly generated data points using same number of classes and appearance clusters, from CIFAR100. }
     \label{fig:tsne9}
 \end{figure}
   \begin{figure}
     \includegraphics[width=.50\textwidth]{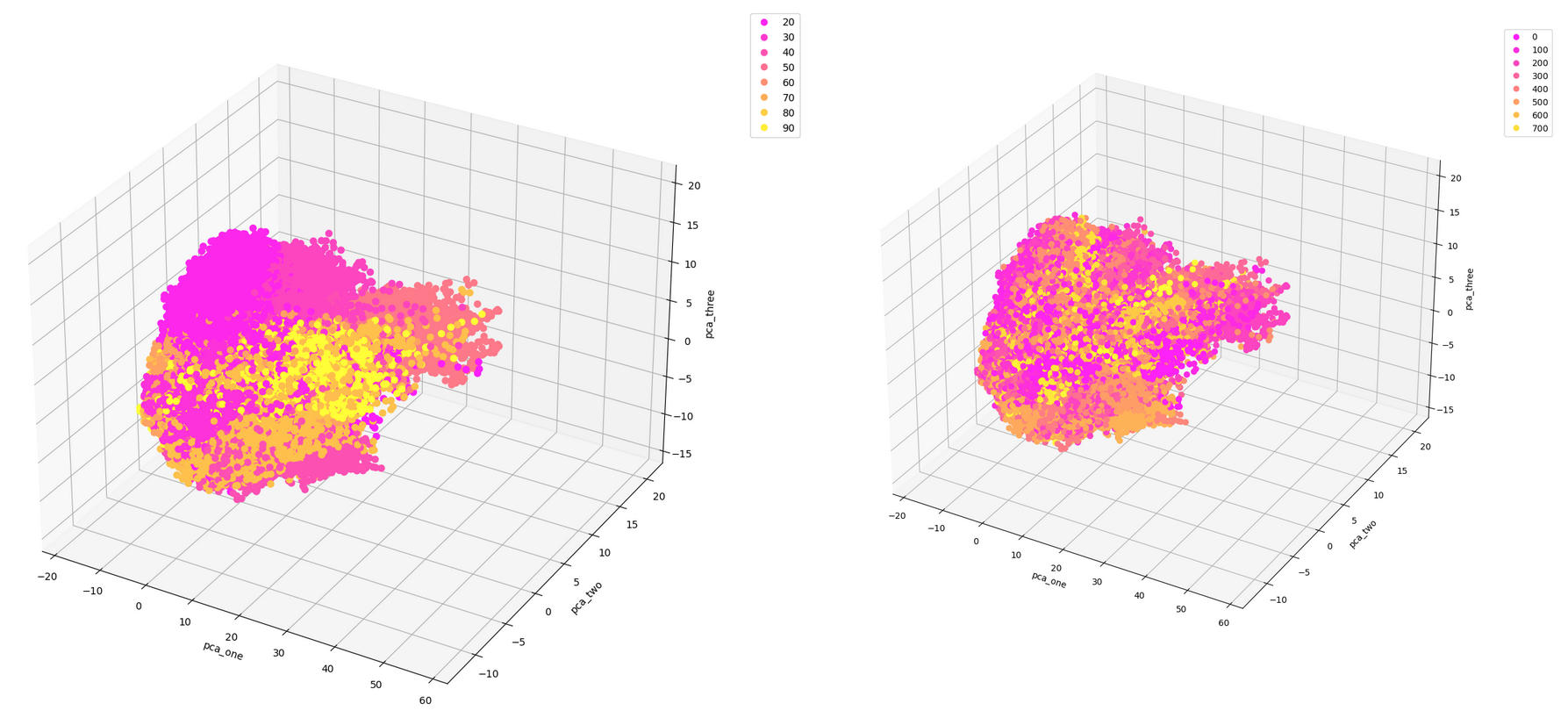}
     \caption{T-SNE feature plots for randomly generated data points using same number of classes and appearance clusters, from CIFAR100. }
     \label{fig:tsne10}
 \end{figure}
   \begin{figure}
     \includegraphics[width=.50\textwidth]{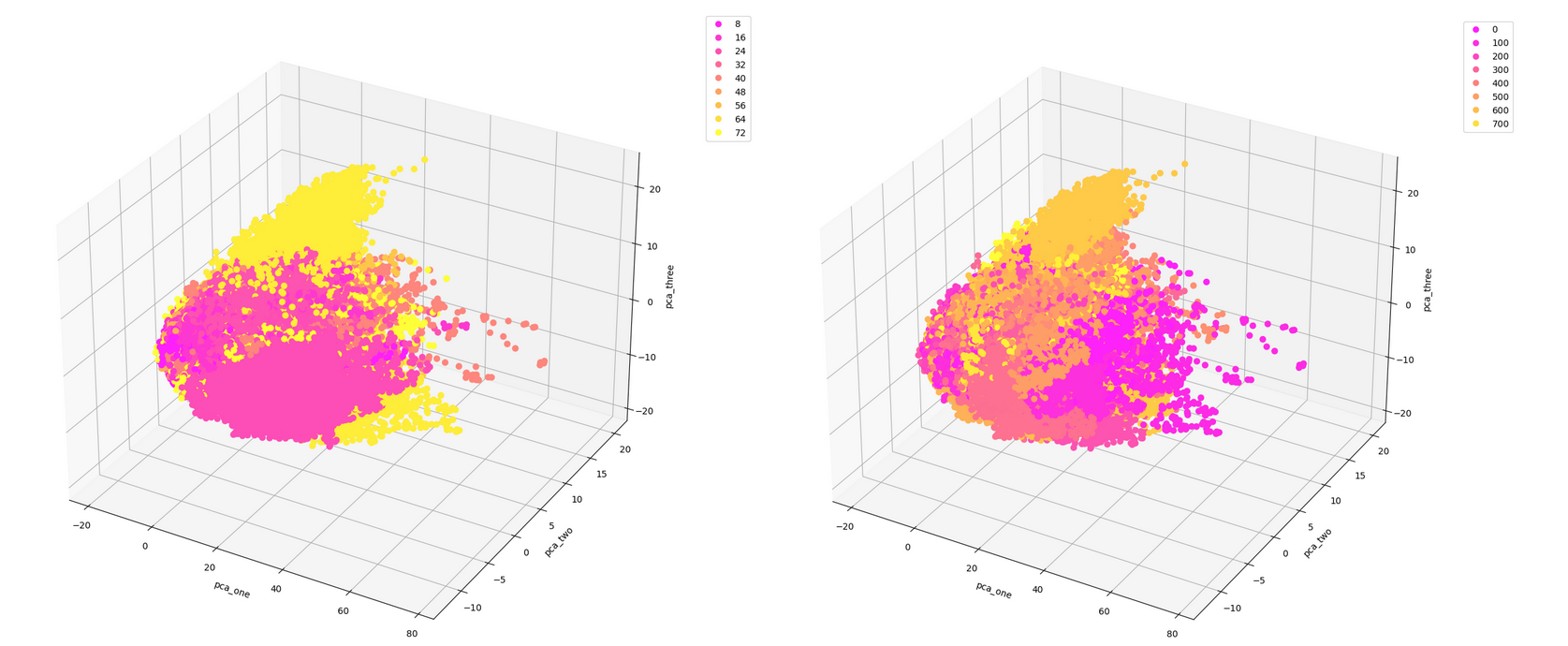}
     \caption{T-SNE feature plots for randomly generated data points using same number of classes and appearance clusters, from CIFAR100. }
     \label{fig:tsne11}
 \end{figure}
   \begin{figure}
     \includegraphics[width=.50\textwidth]{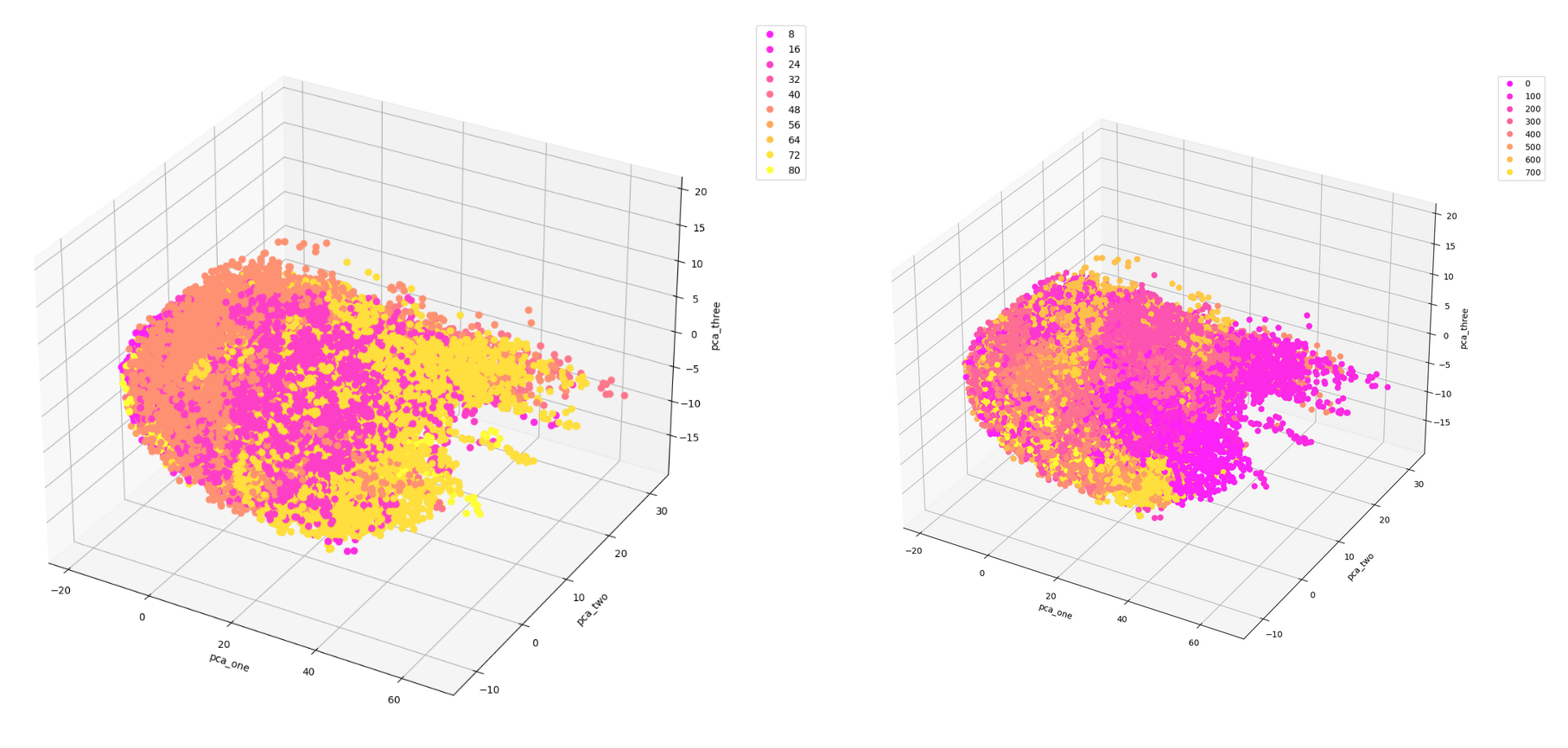}
     \caption{T-SNE feature plots for randomly generated data points using same number of classes and appearance clusters, from CIFAR100. }
     \label{fig:tsne12}
 \end{figure}
   \begin{figure}
     \includegraphics[width=.50\textwidth]{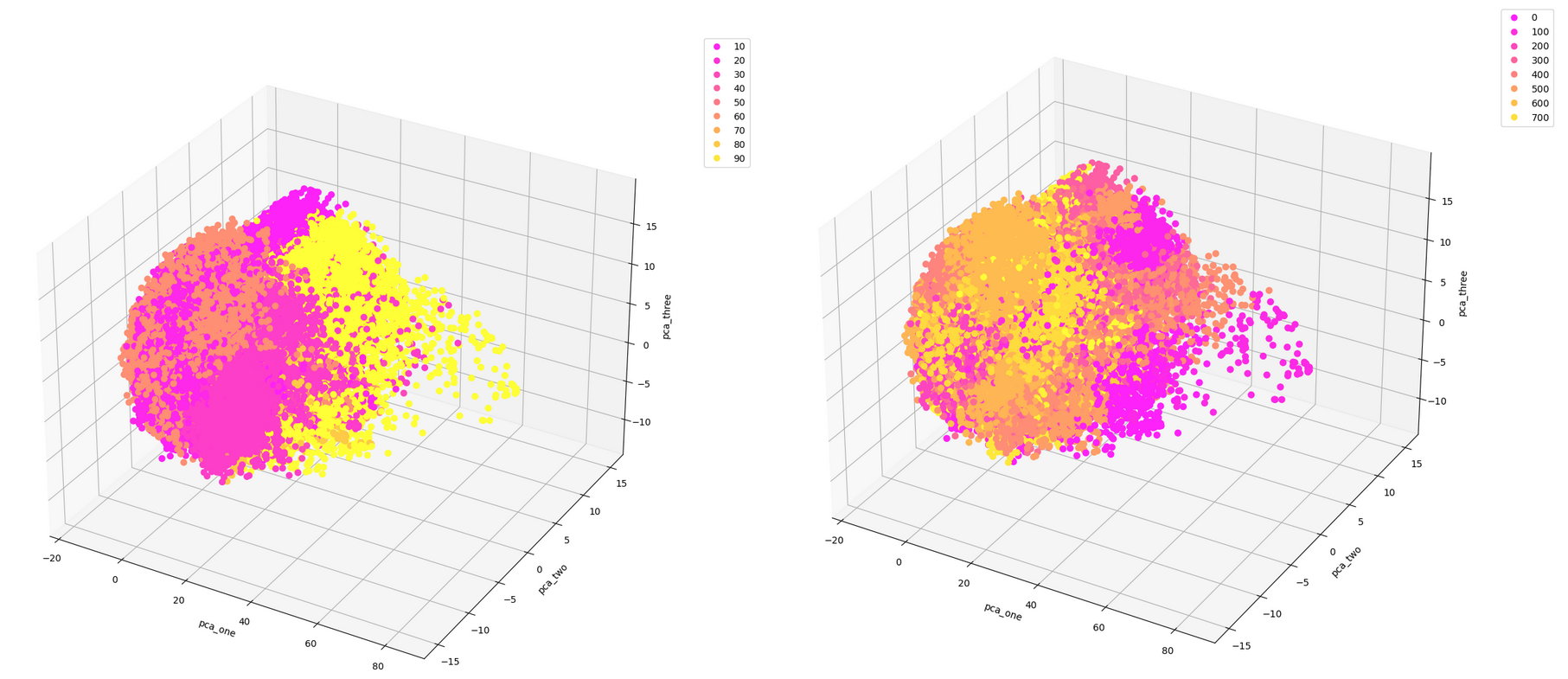}
     \caption{T-SNE feature plots for randomly generated data points using same number of classes and appearance clusters, from CIFAR100. }
     \label{fig:tsne13}
 \end{figure}
   \begin{figure}
     \includegraphics[width=.50\textwidth]{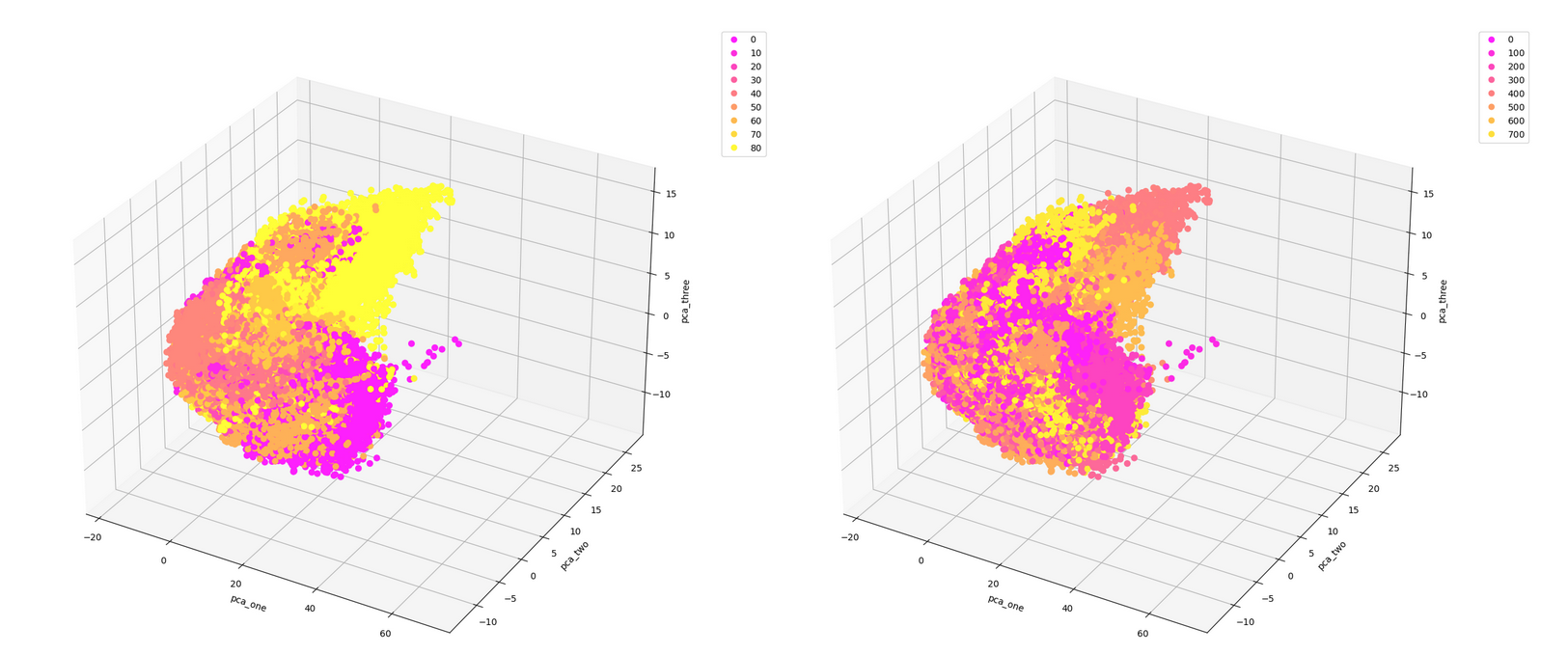}
     \caption{T-SNE feature plots for randomly generated data points using same number of classes and appearance clusters, from CIFAR100. }
     \label{fig:tsne14}
 \end{figure}
   \begin{figure}
     \includegraphics[width=.50\textwidth]{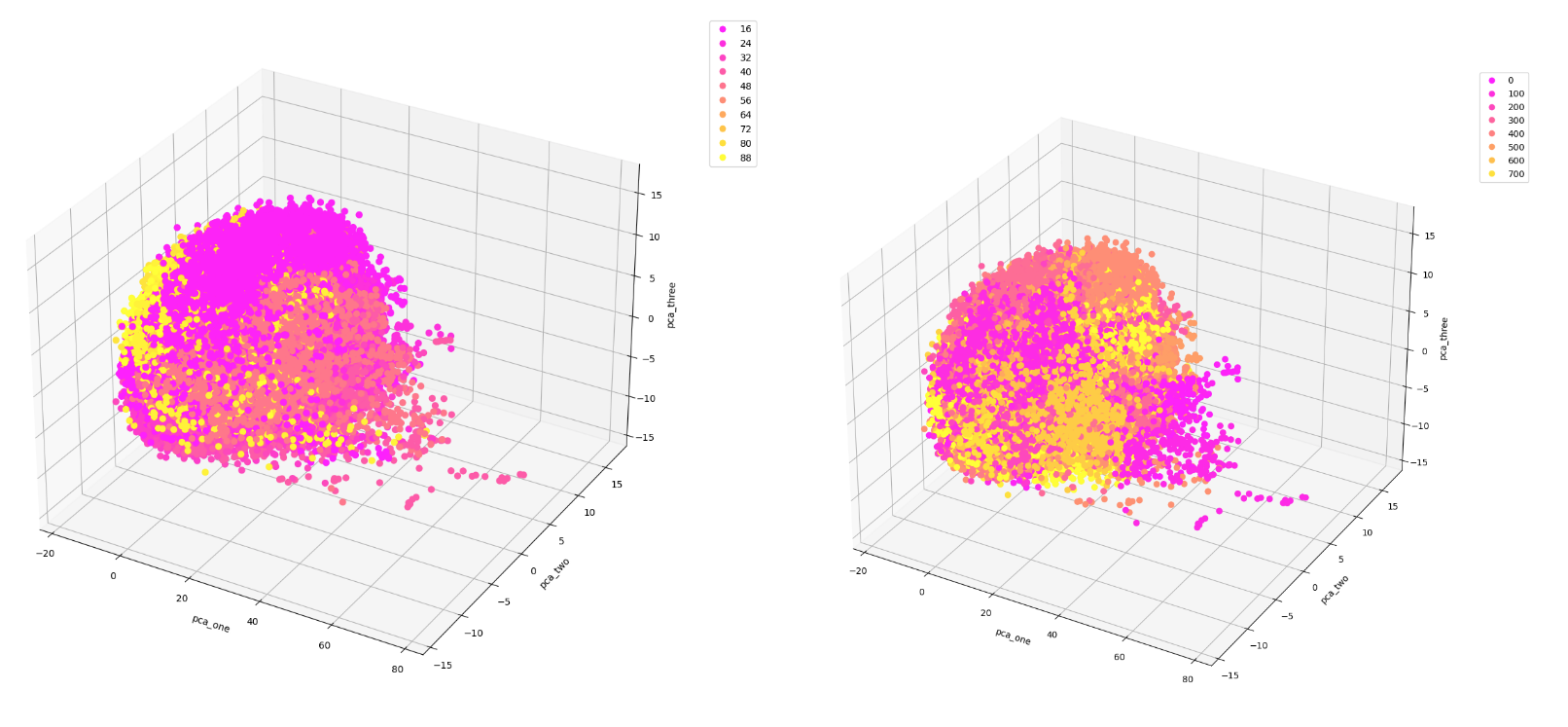}
     \caption{T-SNE feature plots for randomly generated data points using same number of classes and appearance clusters, from CIFAR100. }
     \label{fig:tsne15}
 \end{figure}
   \begin{figure}
     \includegraphics[width=.50\textwidth]{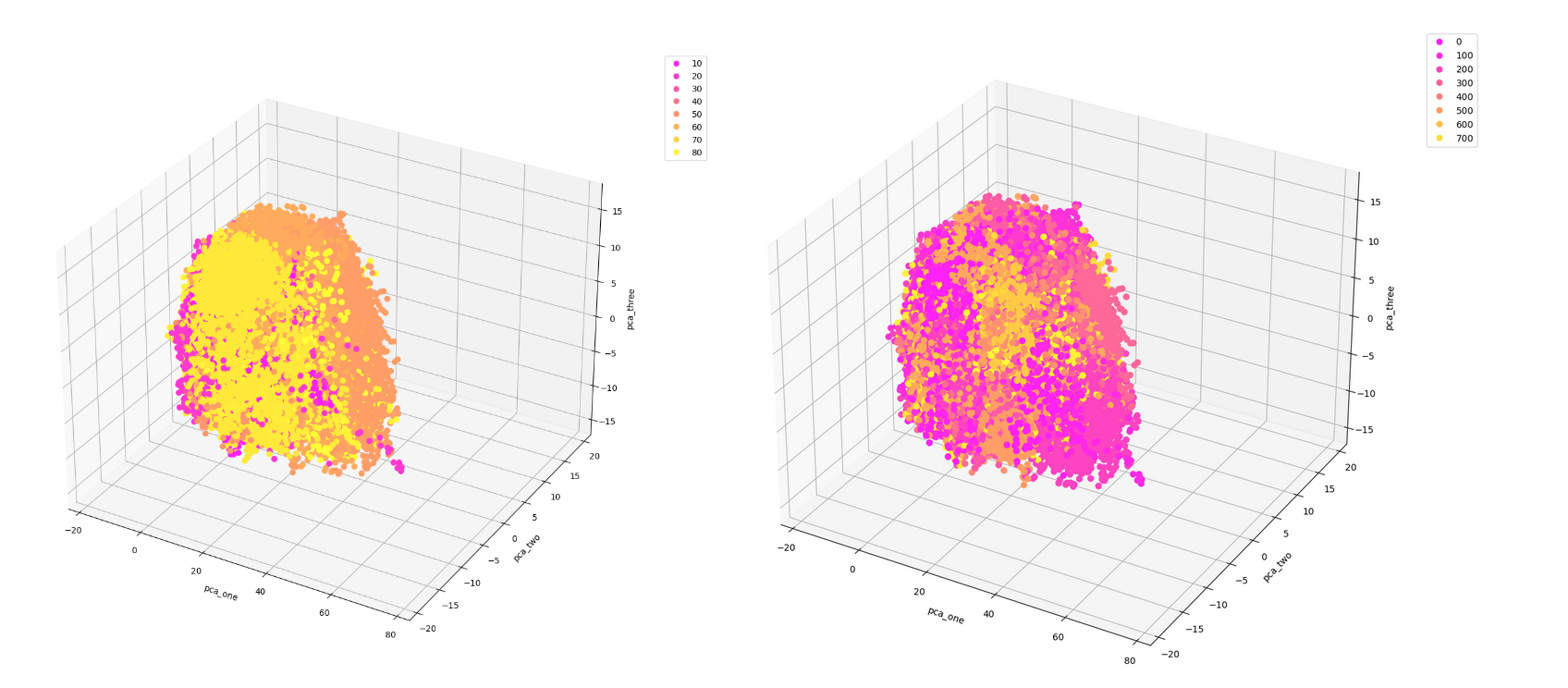}
     \caption{T-SNE feature plots for randomly generated data points using same number of classes and appearance clusters, from CIFAR100. }
     \label{fig:tsne16}
 \end{figure}
 



{\small
\bibliographystyle{ieee_fullname}
\bibliography{egbib}
}